 \documentclass[journal]{IEEEtran}
\usepackage{cite}
\usepackage{textcomp}
\usepackage{dblfloatfix}
\usepackage{amsmath,amssymb,amsfonts}
\usepackage{algorithmic}
\usepackage{graphicx}
\usepackage{textcomp}
\usepackage[T1]{fontenc}
\usepackage[utf8]{inputenc}
\usepackage[ngerman]{babel}
\usepackage[skip=4pt]{caption}
\usepackage{booktabs}
\usepackage{dcolumn}
\usepackage{units}
\usepackage{array}
\makeatletter
\usepackage{booktabs}
\usepackage{graphicx}
\usepackage{multirow}
\usepackage{subfig}
\usepackage{array}
\usepackage{mathtext}
\usepackage{enumitem}
\usepackage{amsmath}
\usepackage{algorithmic}
\usepackage{algorithm}
\usepackage{epsfig}
\usepackage{url}
\usepackage{epstopdf}
\usepackage{cite}
\usepackage{setspace}
\usepackage{xcolor}

\newcommand\MyBox[2]{
  \fbox{\lower0.75cm
    \vbox to 1.7cm{\vfil
      \hbox to 1.7cm{\hfil\parbox{1.4cm}{#1\\#2}\hfil}
      \vfil}%
  }%
}

\usepackage{xcolor}
\usepackage{subfig}

\newcommand{\manas}[1]{\textcolor{blue}{#1}}
\renewcommand{\manas}[1]{#1}

\newcommand{\scope}[1]{\textcolor{teal}{#1}} \renewcommand{\scope}[1]{#1}
\newcolumntype{f}{>{$}l<{$}}
\newcolumntype{n}{l}
\newcolumntype{N}{>{\normal}c}
\newcolumntype{v}[1]{>{\centering\hspace{0pt}}p{#1}}
\newcolumntype{V}[1]{>{\raggedright\hspace{5pt}}P}
\newcolumntype{B}[1]{>{\boldmath\DC@{.}{,}{#1}}l<{\DC@end}}
\newcolumntype{d}[1]{>{\DC@{.}{,}{#1}}l<{\DC@end}}
\newcolumntype{i}[1]{>{\DC@{.}{,}{#1}\mathnormal\bgroup}l<{\egroup\DC@end}}
\newcolumntype{s}[1]{>{\DC@{.}{,}{#1}\mathsf\bgroup}l<{\egroup\DC@end}}
\newcolumntype{R}[1]{%
  >{\begin{turn}{90}\begin{minipage}{#1}\scriptsize\raggedright\hspace{0pt}}l%
  <{\end{minipage}\end{turn}}%
}
\newcolumntype{x}{>{\scriptsize\raggedright\hspace{0pt}}X}
\def \sys {\textit{DeepComfort}}

\def\BibTeX{{\rm B\kern-.05em{\sc i\kern-.025em b}\kern-.08em
    T\kern-.1667em\lower.7ex\hbox{E}\kern-.125emX}}
\begin{document}
%\bibliographystyle{ieeetr}
% \history{Date of publication xxxx 00, 0000, date of current version xxxx 00, 0000.}
% \doi{10.1109/ACCESS.2017.DOI}

%\title{DeepComfort: Multi-Task Learning for Thermal Comfort Prediction of Primary School Students in Winters of Composite Climate}
%\title{Multi-task Learning for Multi-class Thermal Comfort Prediction of Primary School Students}
% \title{Multi-task Learning for Multi-class Thermal Comfort Prediction of School Students}
\title{Multi-task Learning for Concurrent Prediction of Thermal Comfort, Sensation, and Preference}
\author{\IEEEauthorblockN{Betty Lala$^\Phi$, Hamada Rizk$^\dag$, Srikant Manas Kala$^\dag$,   Aya Hagishima$^\Phi$}\\
\IEEEauthorblockA{$^\Phi$ Interdisciplinary Graduate School of Engineering Sciences, Kyushu University, Fukuoka, Japan\\ $^\dag$ Graduate School of Information Science and Technology, Osaka University, Japan\\
Email: lala.betty.919@s.kyushu-u.ac.jp, hamada\_rizk@f-eng.tanta.edu.eg, manas\_kala@ist.osaka-u.ac.jp,\\ ayahagishima@kyudai.jp}}

%\tfootnote{``This work was supported by.....''}

%\markboth
%{Author \headeretal: Preparation of Papers for IEEE TRANSACTIONS and JOURNALS}
%{Author \headeretal: Preparation of Papers for IEEE TRANSACTIONS and JOURNALS}

%\corresp{Corresponding author: Betty Lala (lala.betty.919@s.kyushu-u.ac.jp).}
%%%%%%%%%%%%%%%%%%%%%%%%%%%%%%%%%%%%%%%%%%%%
\maketitle

\begin{abstract}

Indoor thermal comfort immensely impacts the health and performance of occupants. Therefore, researchers and engineers have proposed numerous computational models to estimate thermal comfort (TC). Given the impetus toward energy efficiency, the current focus is on data-driven TC prediction solutions that leverage state-of-the-art machine learning (ML) algorithms. However, an indoor occupant's perception of indoor thermal comfort (TC) is subjective and multi-dimensional. Different aspects of TC are represented by various standard metrics/scales viz., thermal sensation (TSV), thermal comfort (TCV), and thermal preference (TPV). The current ML-based TC prediction solutions adopt the Single-task Learning approach, i.e., \textit{one prediction model per metric}. Consequently, solutions often focus on only one TC metric. Moreover, when several metrics are considered, multiple TC models for a single indoor space lead to conflicting predictions, making real-world deployment infeasible.
This work addresses these problems. With the vision toward energy conservation and real-world application, naturally ventilated primary school classrooms are considered. %A large dataset is created through 
First, month-long field experiments are conducted in 5 schools and 14 classrooms, including 512 unique student participants.
%First, a survey-and-measurement study is conducted in 
%the composite climatic region of north India, 
%in 14 classrooms of 
%5 schools, involving 512 primary school students. 
%Next, the dataset is analyzed for important factors that influence thermal comfort.
%^the impact of factors such as cognition, gender, survey timings, etc., is presented.
%location/architecture of the school/classroom, and day and time of the survey/experiment. 
Further, ``DeepComfort,'' a \textit{Multi-task Learning} inspired deep-learning model is proposed. DeepComfort predicts multiple TC output metrics viz., TSV, TPV, and TCV, simultaneously, through a single model. % trained on the TC data.
%It is validated on the standard ASHRAE II dataset and the %primary student 
%dataset created in this study. 
It demonstrates high F1-scores, Accuracy ($\approx$ 90\%), and generalization capability, when validated on ASHRAE II database and the %primary student 
dataset created in this study.  %in predicting subjective TC perception of primary school students, 
%despite the challenge of illogical responses, and data imbalance.
%The single DeepComfort model demonstrates  
DeepComfort is also shown to outperform 6 popular metric-specific single-task machine learning algorithms.
%, DeepComfort demonstrates improvement by as much as (\%) in TSV, TPV, and TCV, prediction
%and the impact of various factors.
To the best of our knowledge, this work is the first application of Multi-task Learning to TC prediction in classrooms.
%to predict the subjective thermal comfort perceptions of school students.
% comprehension, metabolism, and
% The factors of indoor temperature, humidity, air quality, acoustics and daylight, hence plays a significant role in a students’ well being, health and also contribution to their academic outcome –underlining the importance of understanding the indoor thermal comfort in a classroom.
\end{abstract}


% \begin{highlights}
% \end{highlights}

%\begin{keywords}

Thermal Comfort, Machine Learning, Multi-task Learning, Deep Learning, Classification, Prediction, Students, Classrooms
%\end{keywords}

    \maketitle

%\color{teal}
\section{Introduction}
\label{sec:sample1}
The quality of indoor environment effects health, well-being, and productivity of the residents/occupants. 
%There's enough evidence to show that the people living or working in indoor spaces which lack thermal comfort (either too hot or too cold), are adversely impacted. 
In the absence of adequate indoor thermal comfort, the performance of occupant is likely to deteriorate as their ability to make decisions and/or execute professional tasks depreciates \cite{[75]}. Thus, ensuring satisfactory levels of thermal comfort is necessary and  estimating and/or predicting indoor thermal comfort is an important problem in academia and industry.

The advancements in the domain of machine learning~\cite{goodfellow2016a,murphy2012a} and the continually lowering cost of computational resources, has made it possible to solve complex thermal comfort prediction problems. 
Recent studies show that ML-based models are more precise and accurate ~\cite{Access_review, ML_TC_REVIEW_1, ML_TC_REVIEW_2} in predicting thermal comfort of occupants, as compared to the conventional numerical models such as the Predicted Mean Vote-Percentage of Dissatisfied (PMV-PPD) model \cite{Fanger} and the Adaptive model (AM) \cite{Adaptive}. Further, ML models can be designed and trained for predicting both, individual thermal comfort \cite{ACCESS_TSV_PCM} and group thermal comfort \cite{Access_DNN_ASHRAE}.   

\scope{Further, ML-based predictive solutions are data-driven and more suited to ensure energy efficiency while providing indoor thermal comfort (TC). Typically, buildings are designed to meet the thermal comfort needs of residents in three ways, i.e., through smart Heating, Ventilation And Cooling (HVAC) systems, natural ventilation, and mixed ventilation systems. %A landmark survey shows that 
Though HVAC systems are the most effective in ensuring high TC, their energy consumption is up to 50\% of the energy budget of a building, amounting to a staggering 20\% of the total energy consumption in USA \cite{hvacenergy}. Despite smart HVAC control strategies for energy-conservation \cite{aguilera2019a}, this solution is not ecologically sustainable in the long-term. Moreover, HVACs and mixed systems are not affordable in most developing countries, where the bulk of indoor spaces including classrooms are naturally ventilated \cite{ventilation}. 

From the perspective of energy efficiency and conservation, naturally ventilated building seem ideal \cite{ventilation2}. However, natural ventilation also renders occupants more vulnerable to the temporal changes in the weather, making thermal comfort estimation and prediction a challenging task. This work tries to address this problem.}

\begin{figure}[h]
\centering 
    \includegraphics[width=1\linewidth]{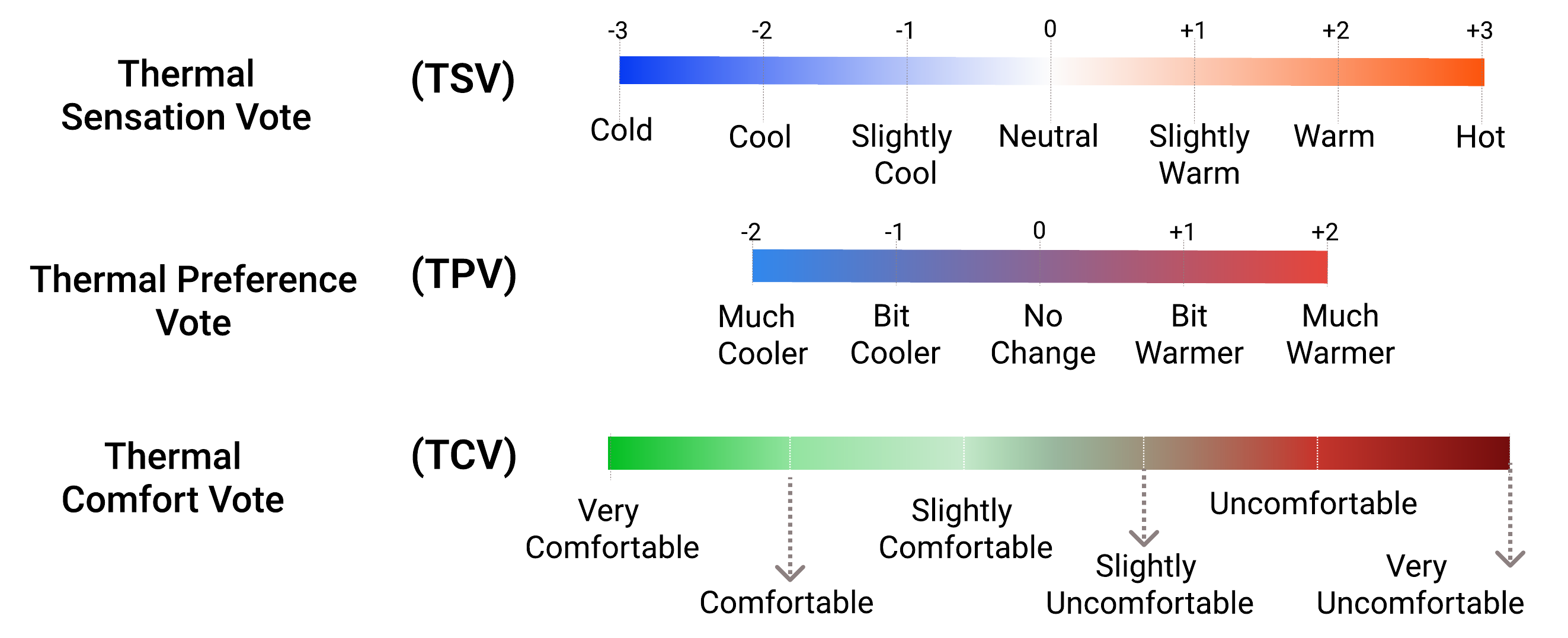}
\caption{Standard Scales for Thermal Comfort Metrics}
\label{fig:tcmetrics}
\end{figure}

 Most importantly, the current ML-based thermal comfort prediction models often offer partial or conflicting solutions. This happens because thermal comfort perception of an individual is highly subjective and personal. It has several dimensions specific to an occupant, such as, \textit{sensation}, \textit{preference}, \textit{current level of comfort}, etc. These dimensions are captured through corresponding subjective metrics, viz., Thermal Sensation Vote (TSV), Thermal Preference Vote (TPV), and Thermal Comfort Vote (TCV), shown in Figure~\ref{fig:tcmetrics}. However, the current ML-based solutions either focus on just one of these metrics, such as TSV \cite{Access_DNN_ASHRAE, ACCESS_TSV_PCM}, or propose a different prediction model for each metric \cite{wang2020b}. This leads to confusing or contradictory predictions, rendering practical real-world implementation infeasible.
 %Such solutions are difficult to implement in a real-world indoor space.
 For practical implementation of TC prediction models, occupant data must be gathered through field experiments. Schools and classrooms are an ideal real-world setting as 
 %as students spend a significant amount of their time confined within their classrooms from kindergarten (age 3-4 years) to university (age 18-30 years). 
school students, spend more time in classrooms as compared to any other built environment outside their homes. It is well established that higher levels of indoor thermal comfort and air quality 
% %of ambient conditions 
facilitate improved concentration and enhance scholastic learning~\cite{[53]}. Moreover, school children exhibit lower metabolism than adults \cite{[59]}, limited cognitive abilities to evaluate their environments, and limited opportunities for adaptation in naturally ventilated classrooms. Thus, understanding and predicting students' thermal comfort needs in naturally ventilated classrooms is a challenging problem that needs to be solved for better learning outcomes, health, and energy efficiency.
 
 %A detailed discussion is presented in the next sub-section.  

To remedy these challenges, this work proposes a Multi-task learning inspired solution, that simultaneously predicts the three most important thermal comfort metrics for primary school students in naturally ventilated classrooms. The high-level research problems and specific contributions are presented ahead.

\subsection{Motivation and Research Problems}
Subjective responses vary across different TC metrics, which makes prediction of occupants' thermal comfort perception a complex problem. The existing TC prediction studies employs a \textit{Single Task Learning} approach, wherein an ML model is dedicated to predicting a single thermal comfort metric. Consequently, studies propose multiple independent models, each focusing on one of the subjective metrics.

However, this approach is problematic. Considering one thermal comfort metric at a time does not adequately capture the thermal comfort of an occupant and may yield conflicting results. For example, the TCV model may predict that the occupant is comfortable (response = ``Comfortable'') while the TPV model may indicate that a preference of a major change in the environment (response=``Much Warmer'').

Further, the pursuit of high accuracy in prediction of a single metric requires fine-tuning of the hyperparameters for that particular TC metric. Thus, a metric-specific model does not guarantee that it will perform reliably for the other metrics. This causes ambiguity and confusion in choosing the \textit{right} TC metric. Moreover, from the perspective of real-world implementation, maintaining and deploying multiple ML models for a single built space is practically infeasible for researchers, building administrators, and indoor residents.

These challenges arise because thermal comfort perception
is subjective and context-specific, making multiple TC
output metrics necessary. One solution is to identify a minimal subset of TC output metrics that have high correlation with all other metrics \cite{wang2020b}. However, this solution requires an additional step of linear and non-linear correlation analysis of only TC output metrics. Doing so may not always yield favorable results leaving the set of TC output metrics unchanged. Even if successful, the process may also result in loss of context-specific information by excluding some of the TC output metrics. 

\scope{The second primary motivation of this work is to pave the way for a real-world implementation of ML-based TC prediction that encourages energy efficiency. TC prediction models are context-specific and the predictive capabilities of a ML-based solution will be sensitive to the characteristics of the occupants, the indoor space, and the outdoor environment. With respect to occupants, thermal comfort prediction of primary school students has been largely unaddressed in the ML-based TC studies \cite{ML_TC_REVIEW_1, ML_TC_REVIEW_2}. Further, naturally ventilated (NV) classrooms are more suitable for energy conservation and long-term sustainability goals. They offer reduced operating costs, lower green-house emissions,  improved indoor air-quality, and prevent the spread of \mbox{COVID-19} \cite{ventilation2, ventilation4}. However,  primary school students are more vulnerable to external environmental factors in NV classrooms. Given their limited cognitive ability to assess their environment and capacity for behavioral adaptation, predicting their thermal comfort responses is far more challenging in NV environments than it is for adults.} 

Thus, this work aims to solve two main research problems with respect to thermal comfort prediction. First, is to address the challenges posed by ``multiple models,'' for the TC metrics in an indoor space. Second, develop an intelligent TC prediction model for primary school students that can be deployed in naturally ventilated classrooms.

\begin{figure*}[htbp]
 \centering%
\begin{tabular}{cc}
    \subfloat[TC Metrics Usage Overall] {\includegraphics[width=.4\linewidth]{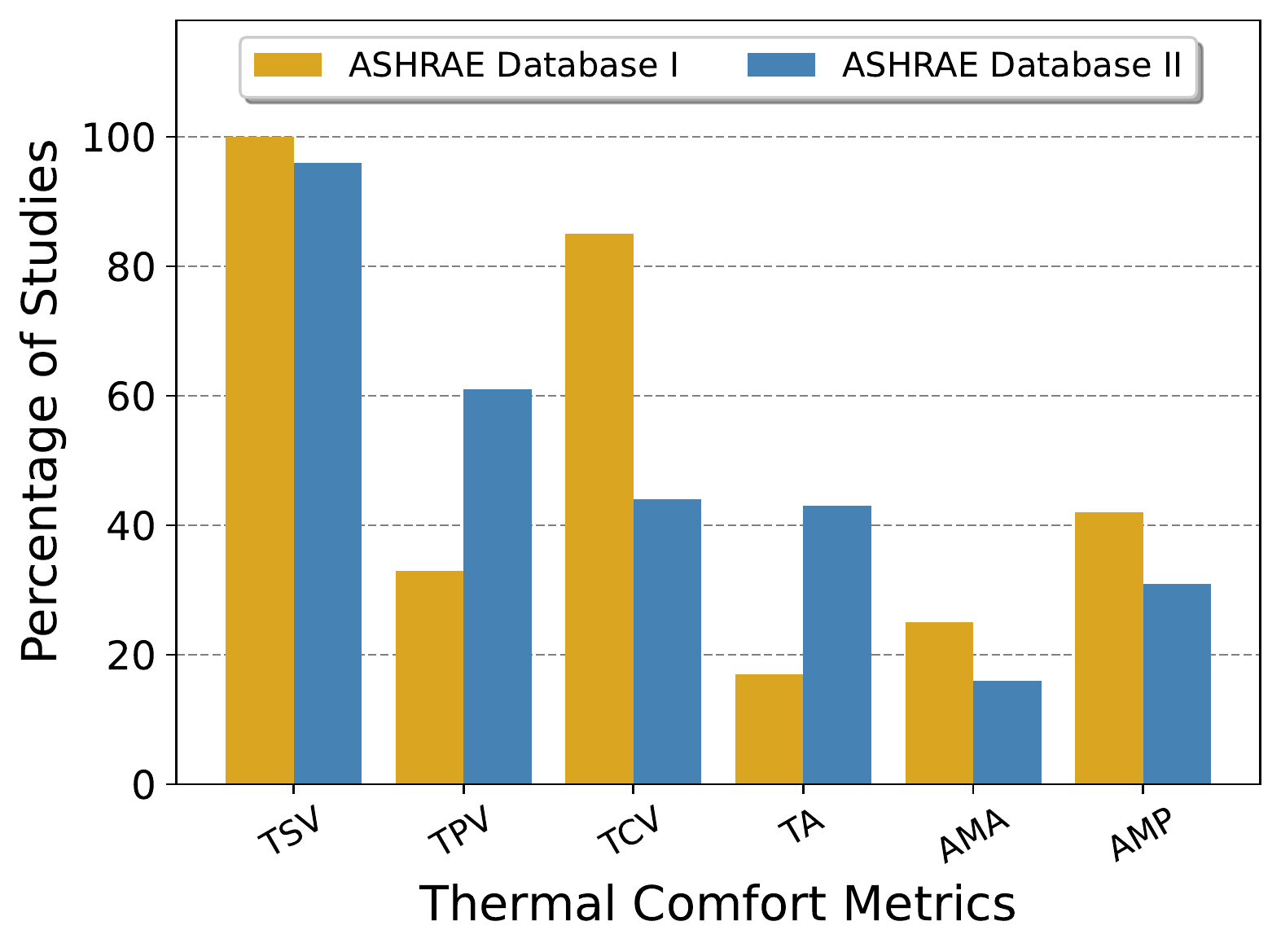}}\hfill\hspace*{0.1cm}%
	\subfloat[TC Metrics Usage in Classroom Studies]{\includegraphics[width=.4\linewidth]{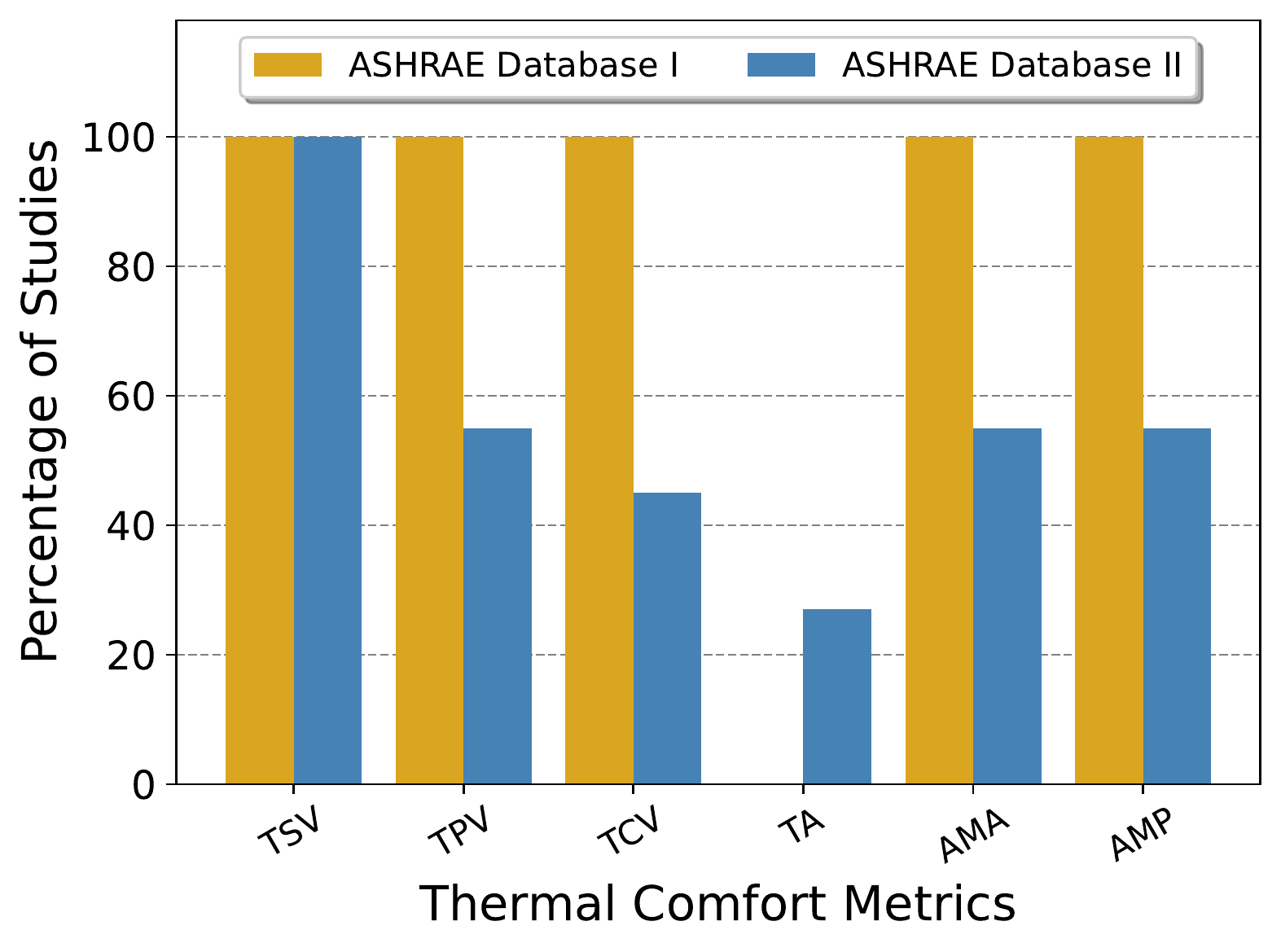}}\hfill
\end{tabular}
    %\vspace*{0.1cm}
  \caption{Use of popular Thermal Comfort Metrics as Outputs in ASHRAE Databases I and II} 
    \label{fig:AshTCmetrics}
    %\vspace*{-0.4cm}
\end{figure*}
\subsection{Contributions}
This paper addresses the above problems by leveraging the \textit{Multi-task Learning} (MTL) paradigm. It adopts a unified approach to thermal comfort prediction through multi-task prediction wherein multiple TC output metrics are predicted simultaneously, with high accuracy, by training a single model. 
%This work 
Further, for accurate context-specific prediction of  thermal comfort perception of primary school students in naturally ventilated classrooms, extensive field experiments and surveys are conducted. 
In particular, the following are the major contributions of this work.
 
 \begin{itemize}
     \item Field-experiments and Surveys: The primary school student dataset is created through month-long field experiments and surveys involving 512 unique student participants, in 14 classrooms of 5 different schools, during January (coldest winter month) in North India. A total of 2039 survey responses were gathered. \footnote{Data relevant to the study is made available to the reviewers here: https://bit.ly/3KTjrp8. The combined winter and summer data will be publicly released after compilation.} 

     \item Data analysis: Distribution of three subjective TC perception responses is analyzed.

     \item Multi-task learning for Thermal Comfort prediction: The work proposes ``DeepComfort'' -- an MTL system, which employs Deep Learning (DL) for accurate and simultaneous multi-class prediction of TSV, TPV, and TCV. \footnote{Model implementation will be made publicly available during final submission.}
     \item Validation: DeepComfort is validated on the standard ASHRAE II dataset and primary student dataset presented in this study. 
     \item Performance Evaluation: DeepComfort is evaluated against 6 single-task machine learning techniques and is shown to outperform them on parameters such as F-score, Precision, Recall, and Accuracy. The STL techniques include supervised shallow algorithms viz., Support Vector Machines, Random Forest,  Decision Tree, K-Nearest Neighbours, Adaboost and unsupervised Deep Neural Networks. 
     \item Impact of categorical features: The multi-task prediction capability of DeepComfort is assessed for different sub-categories of the data viz., gender of the students, grade of students, different schools, field experiment timings, etc. 

  \end{itemize}
To the best of our knowledge, this is the first multi-task learning based thermal comfort assessment study in naturally ventilated classrooms \cite{ML_TC_REVIEW_1, ML_TC_REVIEW_2, ML_TC_REVIEW_3}.

\begin{figure*}[htbp]
\centering 
    \includegraphics[width=1\linewidth]{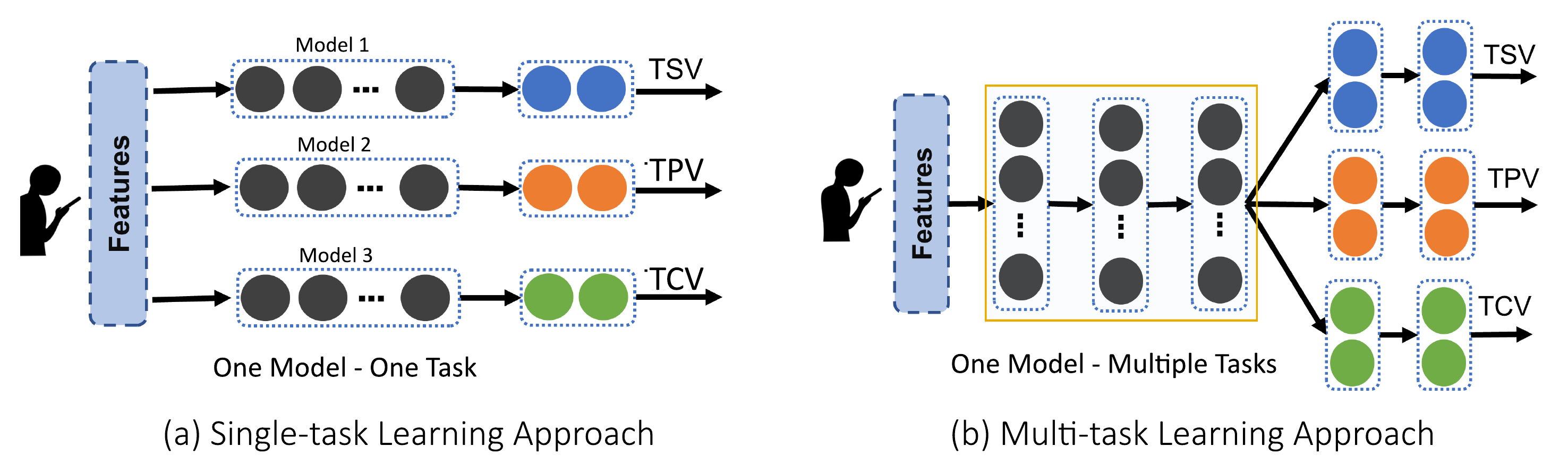}
\caption{Conceptual Schema of Single-task \& Multi-task Learning Paradigms}
\label{fig:stlvsmtl}
\end{figure*}
\subsection{Paper Organization}
The rest of the paper is organized as follows. Section~\ref{sec:review} presents a brief overview of the application of machine learning to thermal comfort studies and discusses the need for multi-task learning. Section~\ref{sec:survey}, describes several aspects of the data gathering exercise in great detail, including the questionnaire, school survey, experiments, and the weather data. Section~\ref{sec:impact} presents an exploratory analysis of important features and TC output metrics.
%the impact of factors such as survey timings, building design, and gender. 
The DeepComfort system is proposed in Section~\ref{sec:system} and the technical details of the underlying Deep Learning neural network model are specified. Thereafter, a comprehensive evaluation of DeepComfort is conducted in Section~\ref{sec:evaluation} along with statistical analysis of data wherever necessary. Finally, conclusions of the study and next-steps are presented in Section~\ref{sec:conclusions}.

\section{Machine Learning for TC Prediction} \label{sec:review}
This section presents a discussion on inputs, objectives, outputs, for ML-based TC prediction models, followed by the limitations of single-task approaches, and the need for multi-task learning.  

\subsection{Input Parameters, Objectives, and Outputs}
%Thesis 
Most machine learning based thermal comfort (MLTC) studies consider features/parameters that are a combination of indoor environmental measurements (\emph{e.g.,} indoor temperature), outdoor environmental measurements (\emph{e.g.,} daily rainfall), and individual features (\emph{e.g.,} clothing) \cite{ML_TC_REVIEW_2}.  Hence, the multi-task DeepComfort model includes input features that are a combination of indoor environmental measurements (\emph{e.g.,} indoor temperature, relative humidity), individual specific features (\emph{e.g.,} Clothing value), and weather data procured from Indian Meteorological Department (IMD)\footnote{The immediate weather data has to be purchased from IMD.} for the month in which the field experiment and surveys were conducted. 

With respect to the objectives of ML models, MLTC studies address a wide array of problems pertaining to the thermal comfort of occupants. These include, predicting thermal comforts of individuals and groups \cite{Access_DNN_ASHRAE, ACCESS_TSV_PCM}, optimizing HVAC systems for energy efficiency of buildings \cite{yang2020a}, predicting occupant behavior \emph{e.g.,} opening/closing windows \cite{kim2018b}, \emph{etc}. The primary objective of DeepComfort is to offer a reliable model for group-based multi-output thermal comfort prediction.
In addition, this work also analyzes the impact of factors such as age, grade, gender, spatial and ambient environment (classroom and school) on multi-task thermal comfort prediction. While impact of age and gender on thermal comfort prediction models have been studied earlier in naturally ventilated buildings \cite{chai2020a}, it has not been done for multi-objective models. More importantly, the DeepComfort model aims to overcome the impact of these factors on the accuracy of multi-task prediction.

Coming to output metrics, the subjective metrics used to quantify thermal comfort in the conventional TC studies are illustrated in Figure~\ref{fig:AshTCmetrics}. It is evident that Thermal Sensation Vote (TSV), Thermal Preference Vote (TPV), and Thermal Comfort Vote (TCV), are the three most popular TC metrics \cite{db2}. Less frequent ones include, Air Movement Acceptability (AMA), Air Movement Preference (AMP), and Thermal Accepatbility (TA). Likewise, in ML-based thermal comfort studies, TSV is used as the sole or primary output in close to 50\% of works, with TPV being used in 12\% studies \cite{ML_TC_REVIEW_1}. Consequently, in this work, TSV, TPV, and TCV are considered to be the outputs for DeepComfort. \footnote{The DeepComfort system can be trained to predict more than three TC outputs as well.}

  \begin{figure*}[b]
 \centering%
\begin{tabular}{cc}
    \subfloat {\includegraphics[width=.24\linewidth]{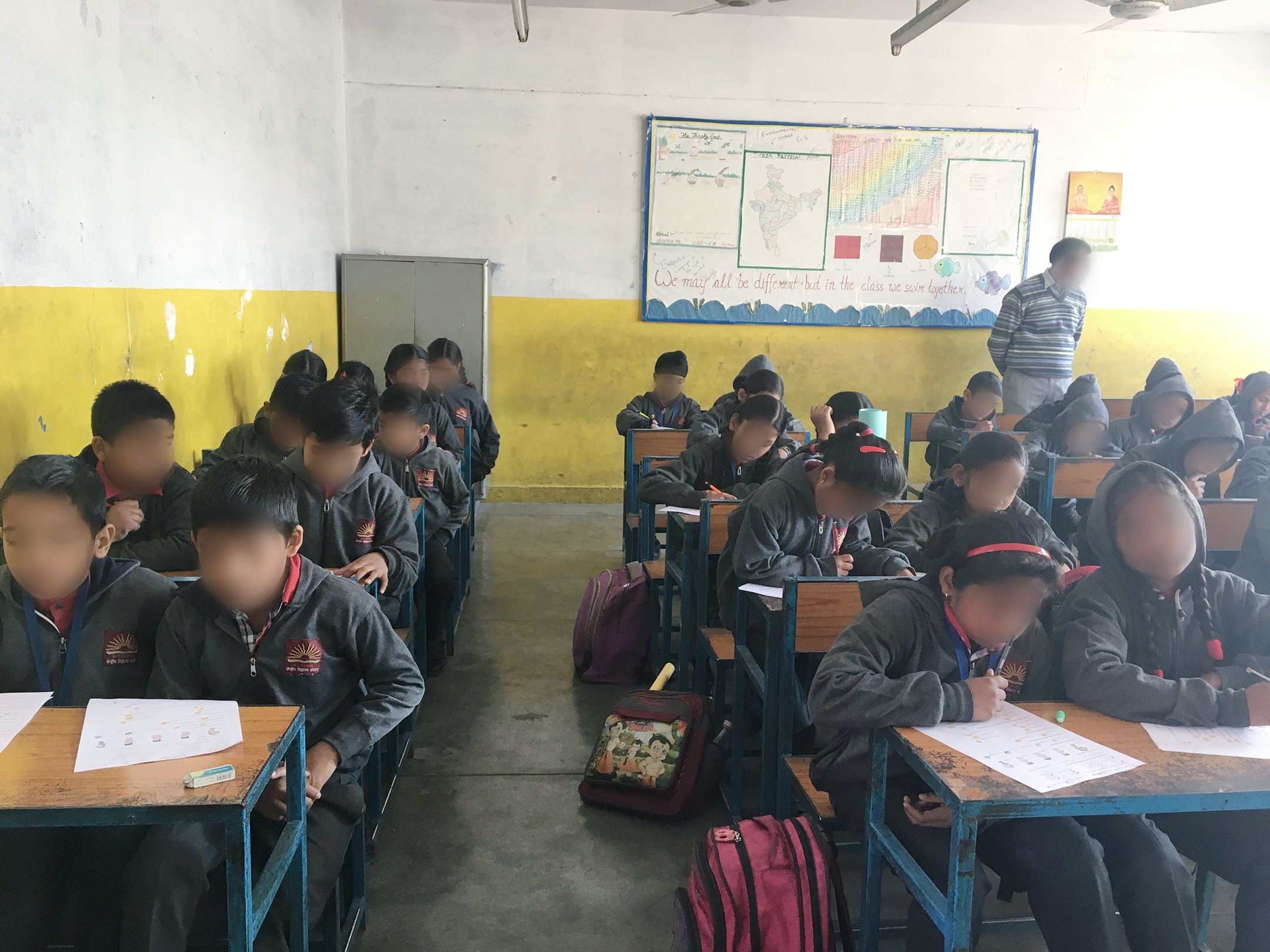}}\hfill\hspace*{0.1cm}%
	\subfloat{\includegraphics[width=.24\linewidth]{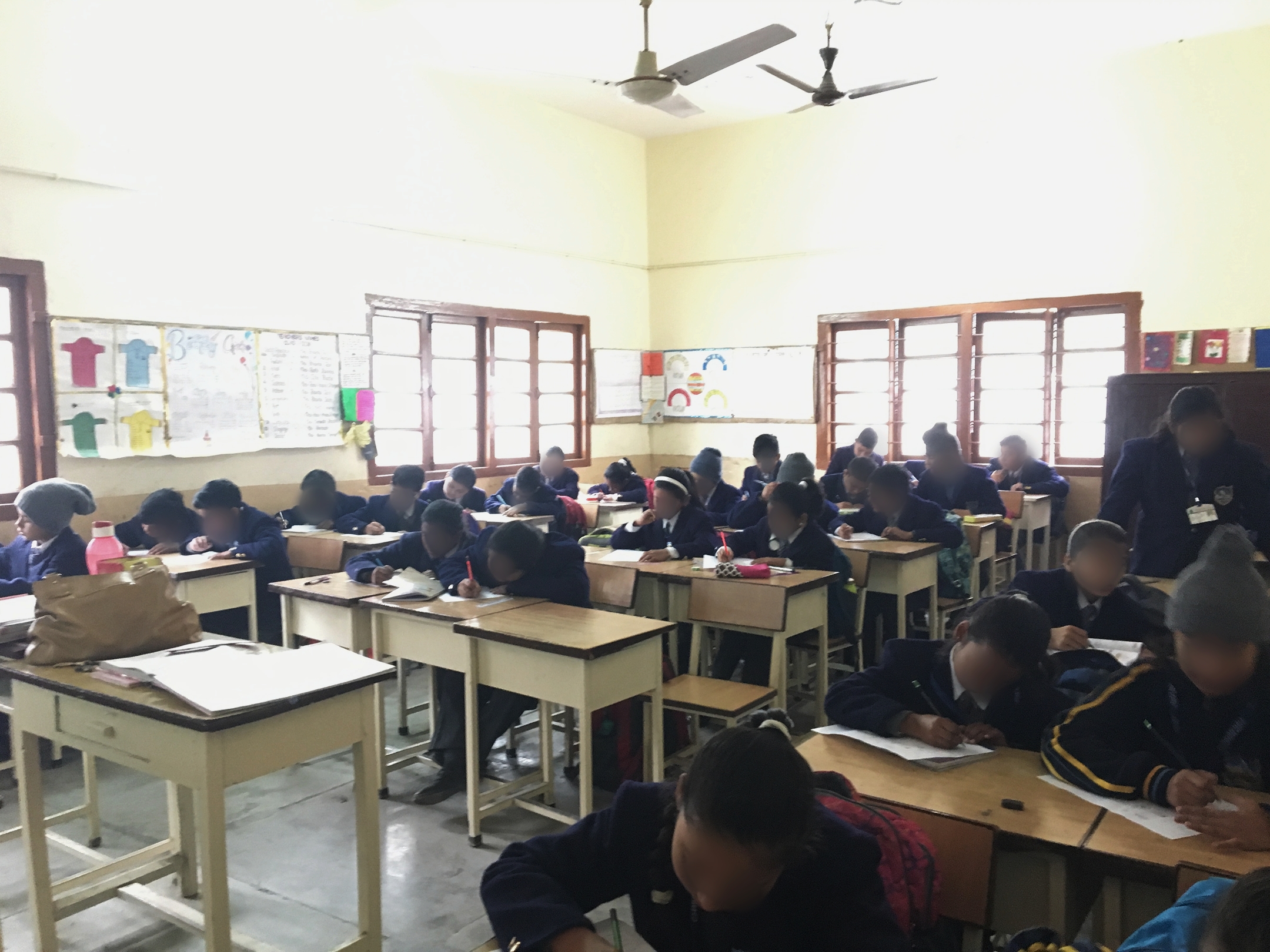}}\hfill\hspace*{0.1cm}%
	  \subfloat{\includegraphics[width=.24\linewidth]{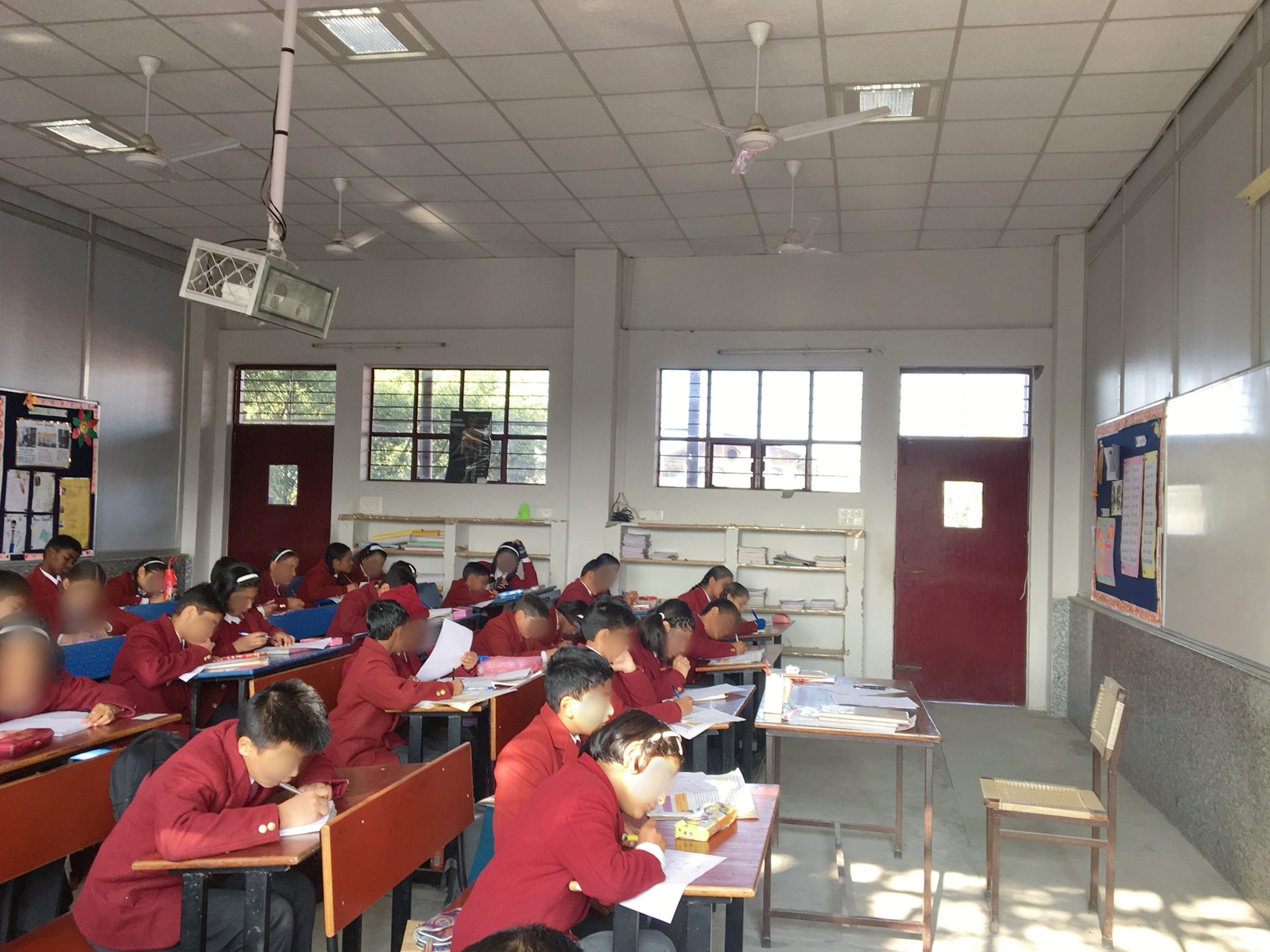}}\hfill\hspace*{0.1cm}%
	\subfloat{\includegraphics[width=.24\linewidth]{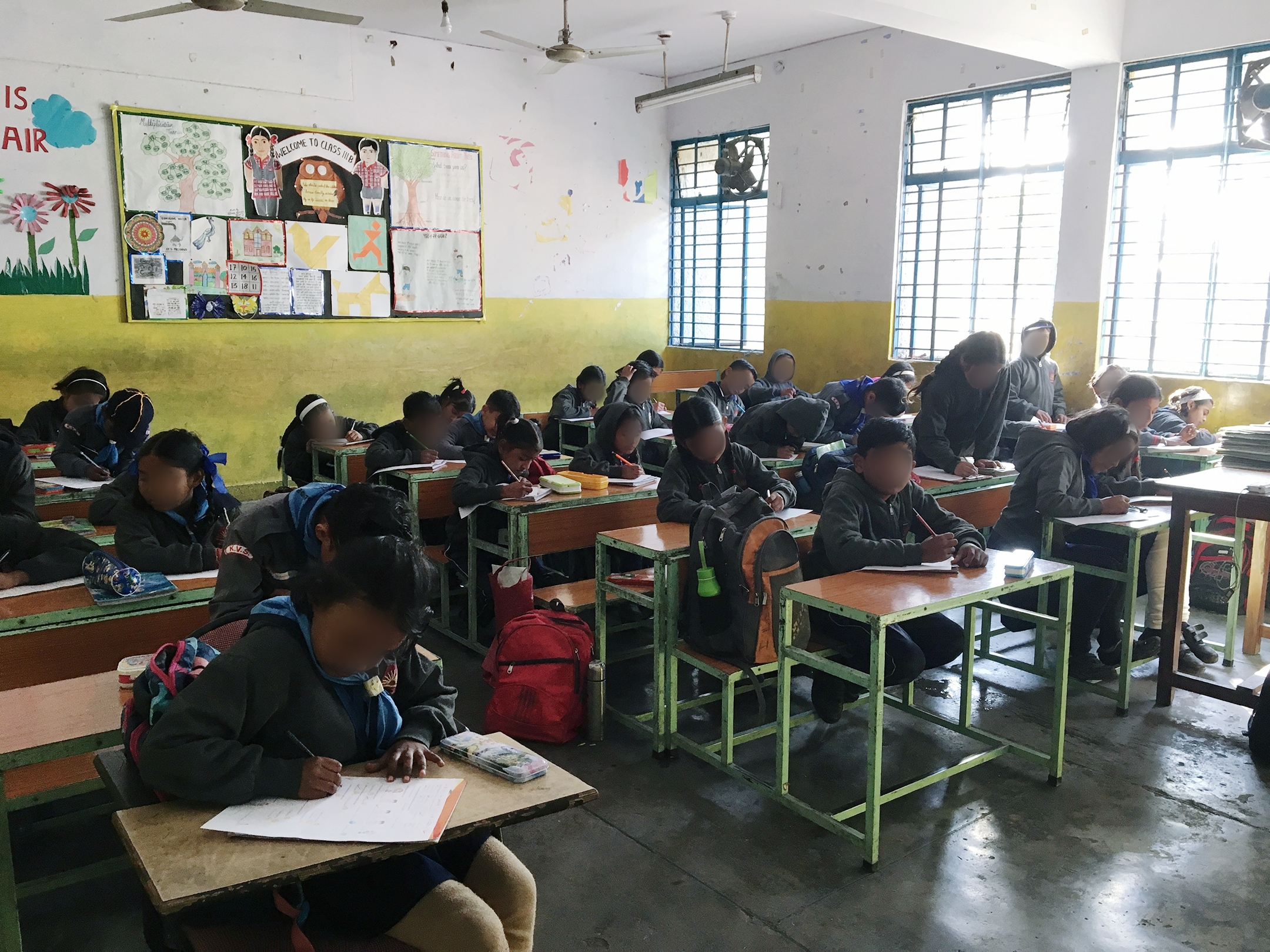}}\hfill
\end{tabular}
    %\vspace*{0.1cm}
  \caption{Survey and Experiments in Classrooms of Different Schools in Dehradun city, India} 
    \label{surveyphotos}
    %\vspace*{-0.4cm}
\end{figure*} 

 \subsection{Single-task \MakeLowercase{vs.} Multi-task Learning}
 \subsubsection{Single-task Learning \& TC Prediction }
  MLTC studies analyze multiple thermal comfort perception metrics and propose ML models that predict these metrics as outputs.  For example, TSV and TCV are used in \cite{chai2020a, katic2020a}, TSV, effective temperature (ET), and standard effective temperature (SET) are used in \cite{wu2018a}, and TSV, TPV, TCV, and TA are used in \cite{wang2020b}.
 However, it is noteworthy that while these studies seek to predict ``multiple'' outputs, they employ \textit{Single-task Learning}, which results in \textit{one ML model per output}. This characteristic is demonstrated as an illustration in Figure~\ref{fig:stlvsmtl}~(a).
 %Thesis It follows that the \textit{proposed ML models are not ``multi-output'' i.e., they do not perform ``multi-task'' learning}. 
 Thus, each output such as TSV, TPV, or TCV, has an independent ML model dedicated to it.
 
The STL thermal comfort prediction models suffer from several problems. First, each output-specific model may differ in the inputs/features required for maximal prediction accuracy.  Further, the number of samples corresponding to different classes of the outputs for example TSV (7 Classes), TPV (5 Classes), and TCV (6 Classes), will vary. It is possible that there may not be sufficient data for each output to train an accurate classification (prediction) model \cite{MTL1}. Specifically, in such scenarios, the models tend to overfit the training data losing its generalization ability for real-world deployment. Moreover, tuning the hyperparameters and optimization techniques such as the number of layers in a neural network model are context-specific and vary across outputs \cite{MTL1, MTL2}. As a result, keeping track of the inputs and model specifications for each output variable is tedious and time-consuming \cite{ML_TC_REVIEW_1}.  
 
 The most critical problem with this approach is that multiple independent models may yield inaccurate and contradicting results. For example, a conflict may occur if the TSV model predicts the sensation felt by occupants to be ``Cold,'' but the TCV model predicts that the occupants are ``Very Comfortable.''  Such contradicting predictions make it challenging to decide which output or model is considered accurate and take corresponding corrective action.   
 
 There are several reasons for conflicting predictions in the single-task learning models. First, the Pearson and Distance correlation analysis of ASHRAE databases I and II presented in \cite{wang2020b} show that all subjective TC outputs do not necessarily exhibit a high correlation with each other. Second, the prediction accuracy of the outputs may depend upon the context, data, and the ML algorithm used. For example, the prediction accuracy for the conventional Support Vector Machine algorithm (SVM) for TPV and TA outputs are 63.9\% and 87.4\%, respectively, leading to a significant performance difference ($\approx$ 36\%) \cite{wang2020b}. Likewise, in \cite{wu2018a}, the prediction accuracy of models does not just vary across outputs viz., TSV, ET, and SET, but also across the choice of the ML algorithm chosen, viz., SVM, Bagging, and Artificial Neural Networks (ANN). 
 Due to these reasons, the single-task approach is not suitable for practical application in a real-world setting.
\begin{table*}[htbp]
    \centering
    \begin{minipage}{0.5\textwidth}
    \caption{Statistical Details of the Survey}
    \centering 
    \includegraphics[width=\linewidth]{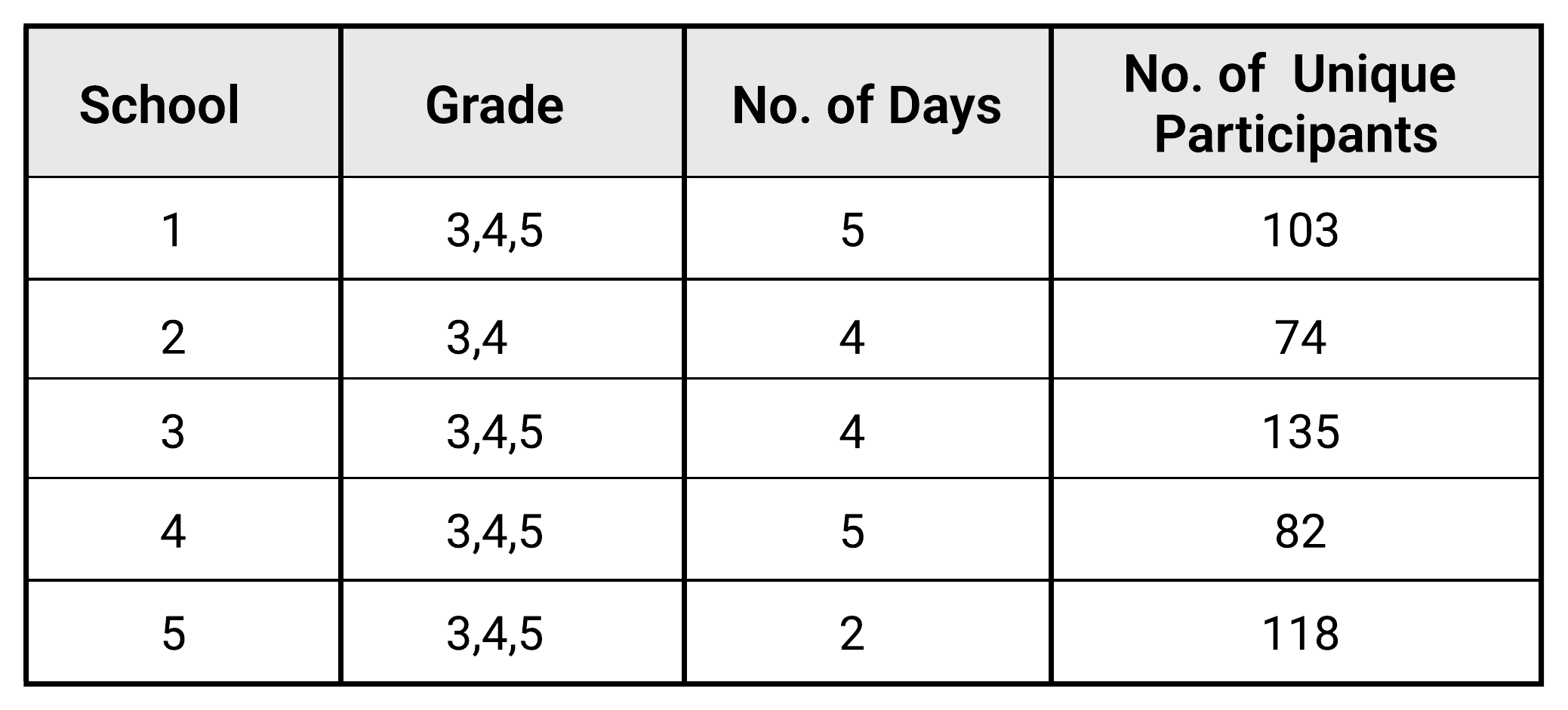}
    \label{surveystats}
    \end{minipage}%
    \begin{minipage}{0.45\textwidth}
    \caption{Details of the Experiment Devices}
    \centering 
    \includegraphics[width=\linewidth]{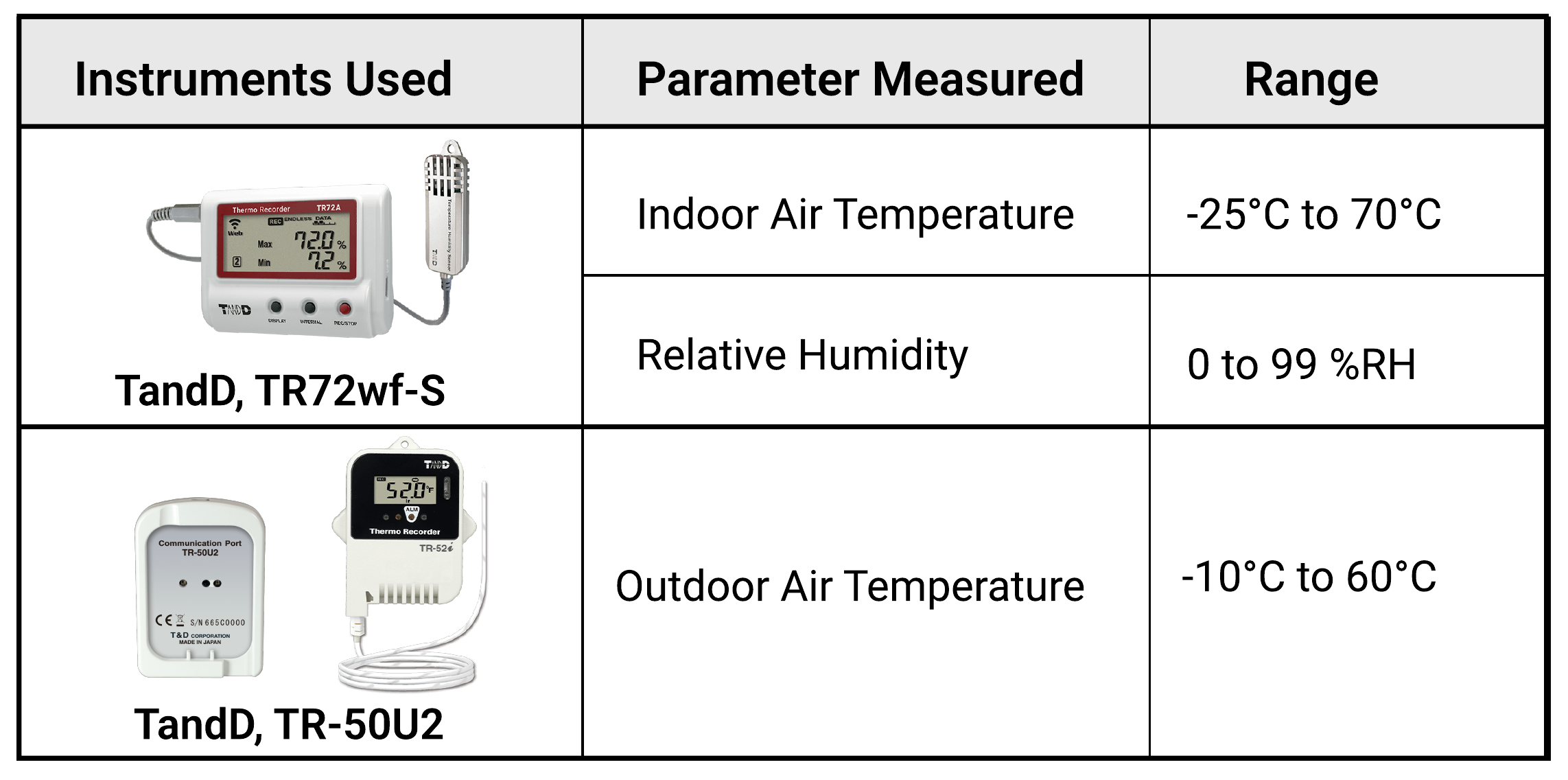}
    \label{devices}
    \end{minipage}
\end{table*}
 \subsubsection{Multi-Task Learning: One Model to Predict them All}
 %\hamada{Hamada san, please make a STL vs MTL diagram if necessary.}
 Multi-task learning (MTL) \cite{caruana} is the solution to the challenges highlighted above, and is illustrated in Figure~\ref{fig:stlvsmtl}~(b). A multi-task model is trained differently in the following respects:
 \begin{enumerate}
     \item It is a common-input-multiple-output system.
     \item The model and hyperparameters for all TC outputs are mainly constant.
     \item It optimizes for the cumulative prediction accuracy of all outputs.
 \end{enumerate}
 Thus, the constancy of input features combined with concurrent learning of multiple outputs gives multi-task learning a clear edge in terms of practical implementation in the real-world. The advantage that MTL offers is best summarized by Caruana et al., ``MTL improves generalization by leveraging the domain-specific information contained in the training signals of related tasks'' \cite{caruana}.
 
%Rewrite the ''MTL'' about the point below.
 The paradigm of Multi-task learning (MTL) has recently been applied to the domain of thermal comfort, primarily to solve the challenges of energy efficiency of buildings and HVAC control \cite{MTL1, MTL2, MTL3, MTL4}. In \cite{MTL2}, authors employ multi-task learning to propose a portable building management solution for better HVAC control. The task-definition is based on publicly available building metadata such as the \textit{Brick} database \cite{Brick} and the solution is validated on the ASHRAE RP884 database \cite{db1}. Using metadata for task-identification is suitable to avoid the problem of negative transfer \emph{i.e.,} incorrect task construction and learning unrelated tasks \cite{MTL3}. However, the use of metadata in MTL is also challenging due to the problems of inaccurate representation generation, need for domain expertise in creating metadata, variation in the context and types of the metadata itself, and improper integration with the MTL system \cite{MTL2, MTL3}. 
 
 Thus, when task information and relation is clear (e.g., optimizing heating and cooling in HVAC) and for specific contexts (e.g., residential buildings), MTL without metadata is equally suitable for thermal comfort prediction. For example, a Deep Reinforcement Learning (DRL) model that aims to optimize the HVAC efficiency with cooling and heating as its two outputs is proposed in \cite{MTL1}. Likewise, the recEnergy system proposed in \cite{MTL4}, leverages a multi-task DRL model to optimize three tasks, \emph{viz.} energy efficiency, occupant comfort, and air quality. 
 
 However, the current studies that leverage MTL studies are aimed at optimizing building and HVAC efficiency, and rely on sensor data and metadata. Further, none of the existing MTL studies have been conducted in naturally ventilated built environments (e.g., classrooms) or train their models using subjective survey and questionnaire data \cite{MTL1, MTL2, MTL3, MTL4}. 

Deep Learning (DL) is increasingly being used for thermal comfort prediction as it offers better accuracy than conventional ML algorithms \cite{Access_DNN_ASHRAE}. DL is also generally more suited for accurate multi-task learning \cite{MTLDL1}. It is better equipped to learn shared representations from interrelated tasks through layer-sharing of multi-task networks \cite{MTLDL3}. Given the capabilities of DL the proposed DeepComfort model employs deep neural-networks.

\par In this work, TSV, TPV, and TCV are considered as the outputs (Labels) of the proposed DeepComfort model. The responses for these metrics along with other data were gathered from the survey and measurement exercise, discussed ahead.
\begin{figure*}[htbp]
\centering 
    \includegraphics[width=\linewidth]{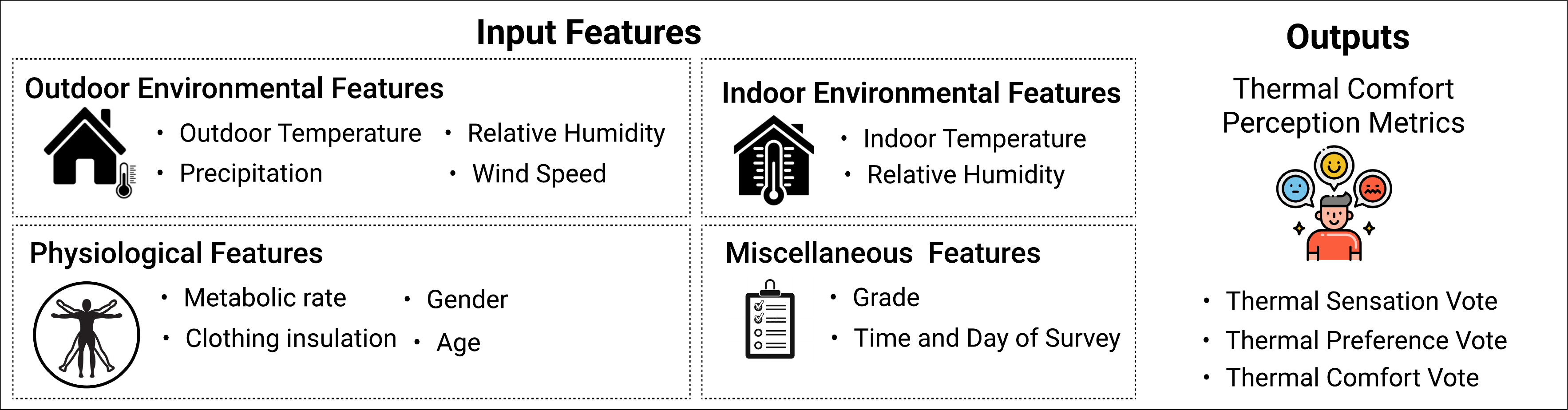}
\caption{Features \& Outputs Considered in the Study}
\label{fig:inout}
\end{figure*}
 \section{Field Experiments and Surveys} \label{sec:survey}
The real-world primary student dataset for the analysis in this work was gathered in the city of Dehradun, which is popularly known as the ``School capital of India.” 
%Both public and private schools with a dedicated primary-school section were considered. 
\subsection{Location and Survey}
The field experiments and surveys were conducted in five schools, namely, Grace Academy, St Thomas School, Kendriya Vidhyalaya, Cambrian Hall and Jaswant Model School. A few photos of the field experiments and surveys are presented in Figure~\ref{surveyphotos}.
To ensure confidentiality, the gathered data is anonymized. 
%so that the conclusions drawn from the school and class specific analysis cannot be traced back to the particular school. 
Henceforth, the schools are denoted as School$_{i}$, where $i\in \{1\ldots5\}$ is randomly assigned to a particular school.

% \begin{table}[htbp]
% \caption{Statistical Details of the Survey}
% \centering 
%     \includegraphics[width=\linewidth]{imgs/table-survey.pdf}

% \label{surveystats}
% \end{table}
% \subsection{Climate and Construction}
% Dehradun city is located in a valley, in the Himalayan state of Uttarakhand, India.  
Dehradun city is located the Himalayan state of Uttarakhand, India, and is characterized by a composite climate, with hot summers and cold winters. Data was gathered for the coldest winter month of January, when temperatures typically fall in the range of 1-20°C. The surveys were administered during school hours on consecutive days, between 8:30 AM--12 PM. It is the coldest period in a working-day and students are likely to experience most discomfort. The typical duration of survey and field experiments in each class/session was 30 minutes.
This study was conducted in 14 naturally ventilated classrooms which makes the thermal comfort prediction more challenging \cite{ML_TC_REVIEW_1, ML_TC_REVIEW_2}. 
Further, the architectural design and construction styles of schools and classrooms considered in this study differ considerably. 
Elements of campus planning for composite climates e.g., open courtyard plan for improved cross-ventilation, can be noticed in the architectural layout of three of the five schools.   

The primary school students from class-levels/grade-levels 3$^{rd}$ to 5$^{th}$, typically belonging to ages 6-13, are the participants of this study. 
\begin{figure*}[htb]
    \centering
    \begin{minipage}{0.51\textwidth}
    \centering 
    \includegraphics[width=\linewidth]{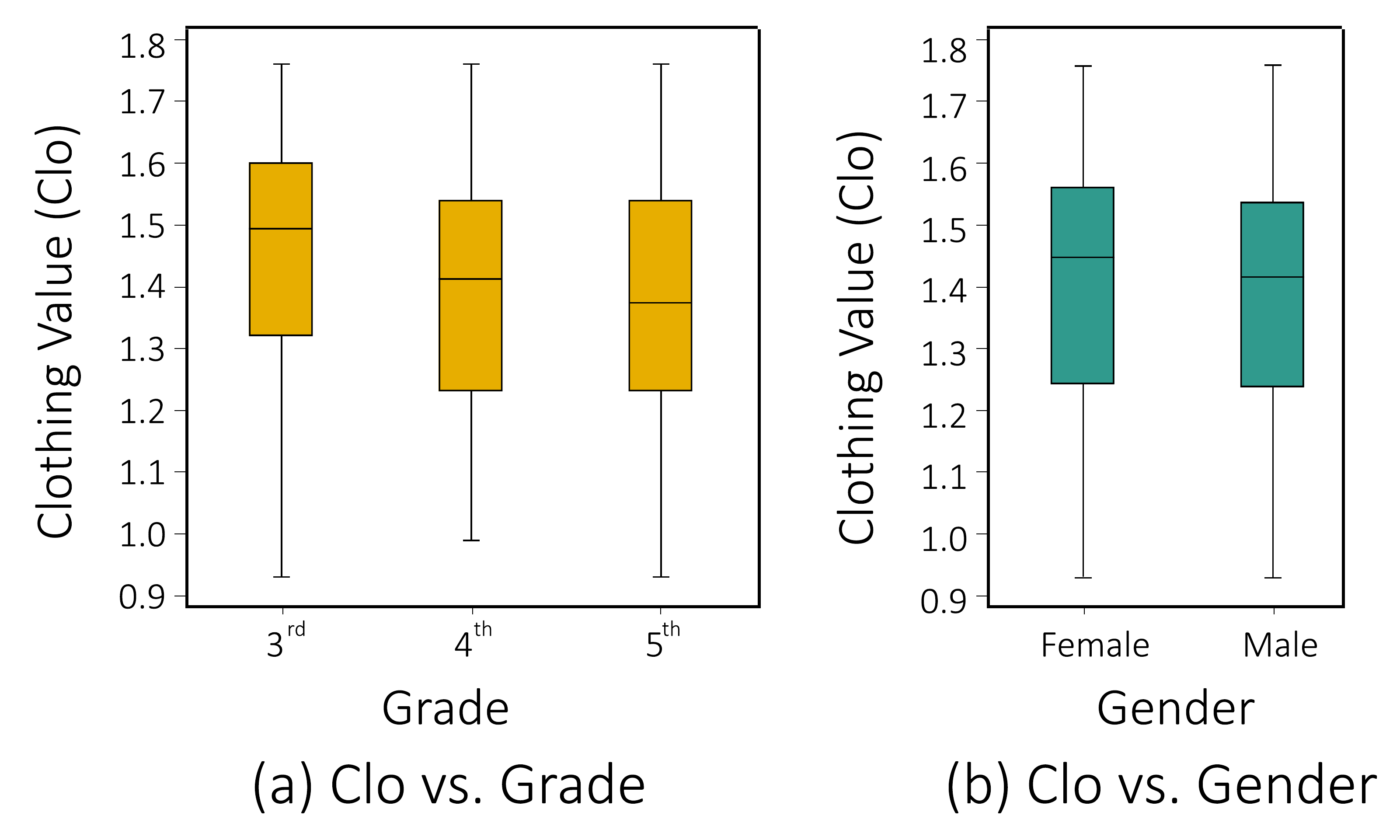}
\caption{Distribution of Clothing Values}
\label{fig:clobox}
    \end{minipage}%
    \hspace{0.1cm}
    \begin{minipage}{0.4\textwidth}
    \centering 
    \includegraphics[width=\linewidth]{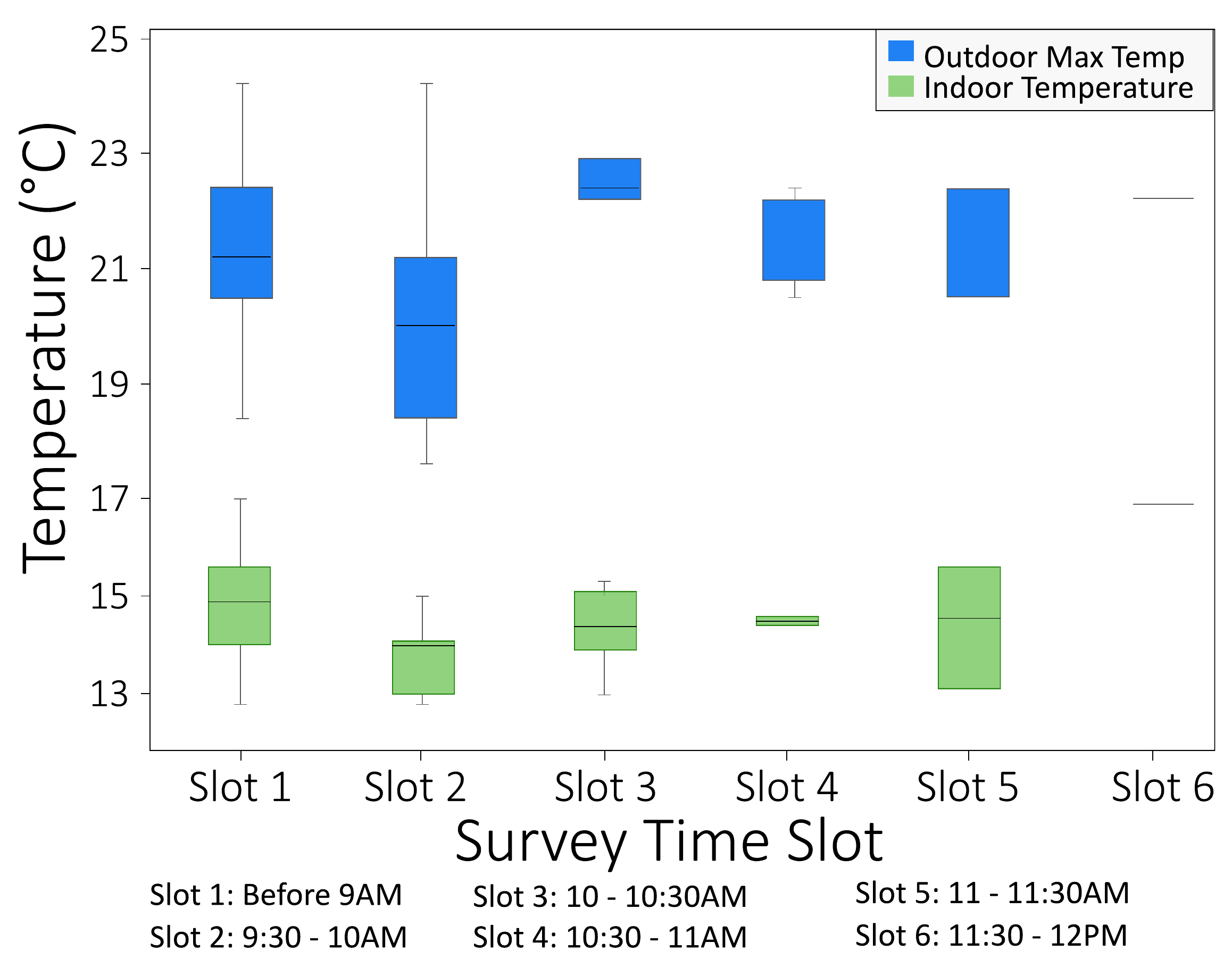}
\caption{Distribution of Temperature Values}
\label{fig:tempbox}
    \end{minipage}
\end{figure*}
Recent surveys highlight the lack of adequate research on ML-based thermal comfort prediction with children as the primary participants \cite{ML_TC_REVIEW_1, ML_TC_REVIEW_2}. This study intends to bridge this gap. The dataset comprises of 2039 responses collected from 512 primary school children as unique participants. A school-wise quantitative distribution of the participants is presented in Table~\ref{surveystats}. 

\subsection{Measurements and Climate Data}
The field experiments are accompanied with the ``right-now" questionnaire, specially designed for primary school students. While the students filled in the questionnaire sheet, parameters of indoor air temperature, relative humidity, and air speed were measured every 2 minutes using \textit{TandD TR72wf-S}. Similarly outdoor air temperature was measured using \textit{TandD TR72wf-S}. The devices were calibrated prior to the measurements and technical specifications are presented in Table~\ref{devices}. Finally, the outdoor weather data for daily maximum, minimum, and average temperatures was collected from Indian Meteorological Department (IMD) of the city for the surveyed days. 
\par
The important features and outputs considered in the analysis and the proposed \sys{} model are depicted in Figure~\ref{fig:inout}. These include measured indoor and outdoor parameters, weather data, and survey data (21 subjective questions). Exploratory data analysis of important features and the three TC output metrics is presented ahead.

\section{Exploratory Data-analysis} \label{sec:impact}
This section seeks
%presents the analysis of important variables in the primary student dataset
to identify relevant patterns in the data and determine the challenges to be expected in the multi-task prediction. In particular, analysis focuses on the variation in clothing (Clo), indoor and outdoor temperatures, and most importantly, the distribution in the three TC output metrics to be predicted concurrently. 

Clothing provides thermal insulation which is vital for thermal comfort, especially in naturally ventilated indoor spaces. Although primary school students who participated in the field experiments usually have a school uniform, it was observed that children added/removed layers of clothing. Further, male and female students are usually prescribed a different dress-code e.g., trousers for males and skirts for females. With this context, two interesting findings with respect to clothing are observed. First, the amount of clothing children wear seems to be reducing with Grade. The mean Clo value for all students in grades 3$^{rd}$, 4$^{th}$, and  5$^{th}$, is 1.375, 1.398, and 1.451, respectively. The pattern can be observed in Figure~\ref{fig:clobox}~(a).  The finding indicates that students are more likely to resort to behavioral adaption, e.g.., modify their clothing, with increase in cognition. The second  aspect, visible in Figure~\ref{fig:clobox}~(b), is that there is a slight variation in average clothing values based on gender. Female students have a slightly higher Clo on average (1.417) than male student participants (1.403).

\begin{figure}[h]
\centering 
    \includegraphics[width=\linewidth]{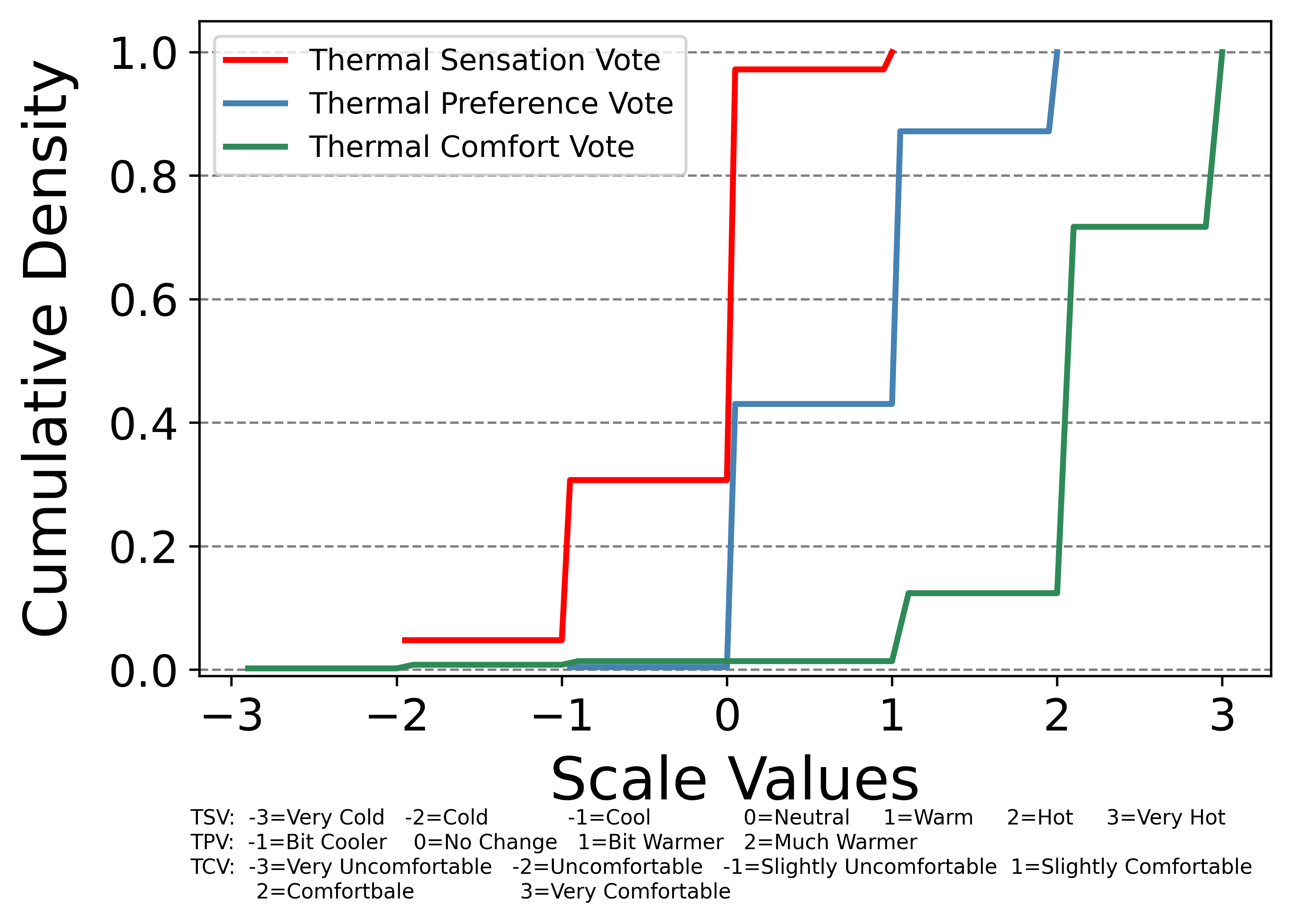}
\caption{Distribution of TSV, TPV, and TCV}
\label{fig:CDF}
\end{figure}

The distribution of indoor temperature during the field experiments and the daily maximum temperature, with respect to survey timings is presented in Figure~\ref{fig:tempbox}.  In general, the median indoor temperature is almost 7°C lower than the median daily temperature. Further, the median indoor temperature stays more or less around 15°C, as compared to median outdoor temperature that ranges approximately in the range of 20°C to 23°C. More importantly, as the day progresses, the outdoor temperature stabilizes, and shows little variability. e.g., Slot~4 variance, for outdoor maximum temperature and indoor temperature, is 0.748 and 2.468, respectively. This clearly effects the TC perception of occupants, thereby influencing a prediction accuracy of classification models.

%\subsection{Need for Multi-task Learning}
A high-level comparative analysis of the distribution of TSV, TPV, and TCV responses is presented in Figure~\ref{fig:CDF}. The empirical cumulative distribution reveals the complexity in thermal comfort prediction due to the use of multiple subjective responses. A smaller percentage of student responded to feeling a ``Cool'' or ``Cold'' sensation (TSV=-1 or -2), while a much larger proportion of student responses indicate that they prefer the classroom environment to be ``Bit Warmer'' or ``Much Warmer'' (TPV=1 or 2). What makes the problem more challenging is that the sensation and preference indicated by the students is not reflected in their comfort votes. Most students (regardless of the school or time-slot) claim to be experiencing varying degrees of comfort (TCV=1 or 2 or 3), which contradicts the TSV and TPV trends. This indicates a small but significant volume of ``illogical responses,''  as it can impact the accuracy (precision) of multi-class classification models. This problem  highlights why the \textit{one model per metric} approach to predict occupants' thermal comfort is not desirable. 

The paradigm of Multi-task Learning offers efficient and practical solutions to these challenges. The following section discusses the proposed \sys{} MTL system in great detail. 

\section{System Vision and Implementation}  \label{sec:system}
This study envisions a practical and feasible real-world implementation of a thermal comfort prediction model. The comprehensive large-scale survey was conducted with that vision in mind. However, implementing the proposed \sys{} model in classrooms will require resources and greater institutional participation. Nevertheless a high-level overview of practical solution is presented next, followed by the technical details of the implementation.  
\begin{figure}[H]
\centering 
    \includegraphics[width=1\linewidth]{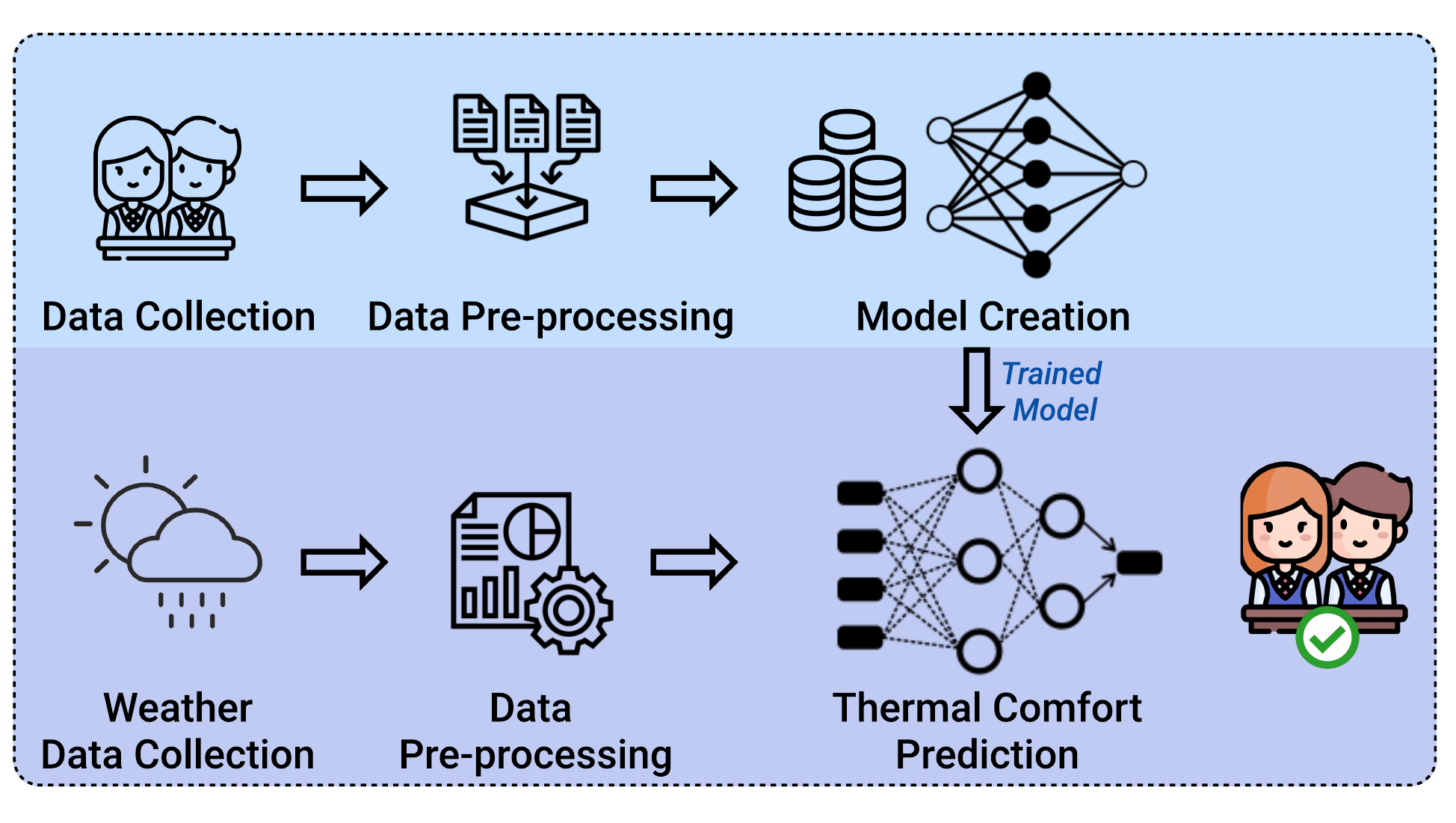}
\caption{The \sys{} System Architecture}
\label{fig:system}
\end{figure}
\begin{figure*}[h]
\centering 
    \includegraphics[width=\linewidth]{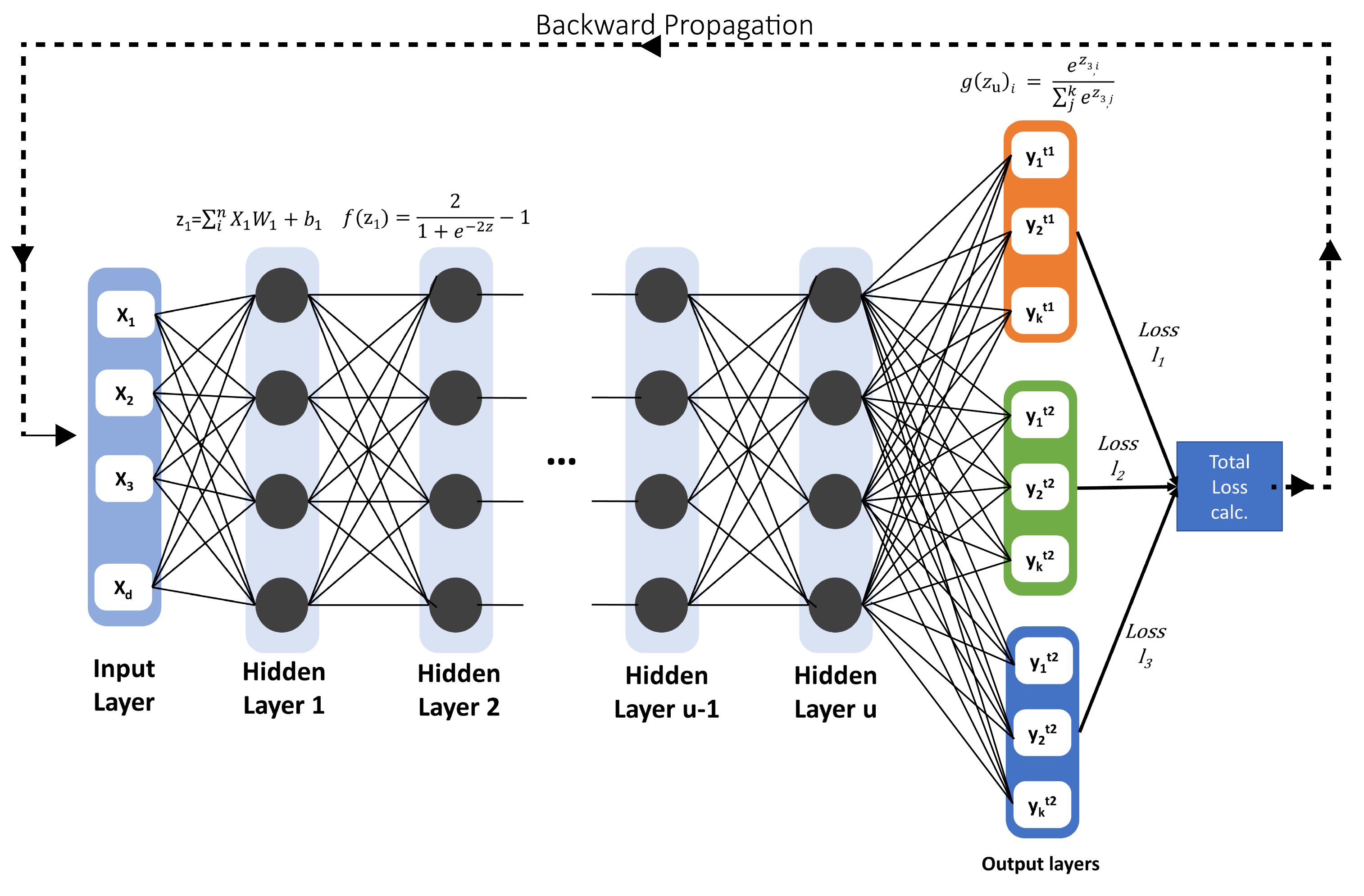}
\caption{The Proposed Multi-task Deep Neural Network Structure}
\label{fig:network}
\end{figure*}

\subsection{System Concept and Vision}
The \sys{} system architecture in illustrated in Figure~\ref{fig:system}. It is conceptualized as a two-stage system with an offline data gathering and training stage and an online thermal comfort prediction stage. \sys{} initializes the offline stage by data gathering of ambient temperature, relative humidity, clothing level, etc.,  and students' subjective thermal comfort responses.  
The data can typically be collected by a Thermal Collector App running on the student's tablets in school. In this study, the data was gathered using paper questionnaires at schools as tablets could not be used due to logistical reasons. Thereafter, the collected data is sent to the local server
% in the cloud 
for further processing. The pre-processing modules handle missing information and put the data in a format appropriate for further processing. 

Next, the Model Creation module builds and trains a deep learning-based multi-task learning model to accurately predict students' thermal comfort, thermal sensation, and thermal preference. This module also determines the optimal model hyperparameters (discussed in Section~\ref{sec:parameters}). To do so, the module employs \textit{grid search} -- an exhaustive search which iterates on numerous combinations of parameters' values until the optimal value, that maximizes the model accuracy, is achieved.

Finally, the trained MTL model is stored for later use in the \textit{online phase}. During the online phase, the school admins or instructors can at any time estimate the thermal comfort of each student by providing the inputs (e.g., ambient temperature, clothing values etc) to the trained model in the offline stage. 

\color{black}

%\subsection{The DeepComfort Model}  \label{sec:system}

% Table \ref{table:notations} summarizes the notations used in this section.

% \subsection{Data Collection}  \label{sec:data_colelction}

\begin{figure*}[htbp]
 \centering%
\begin{tabular}{cc}
    \subfloat[TSV] {\includegraphics[width=.33\linewidth]{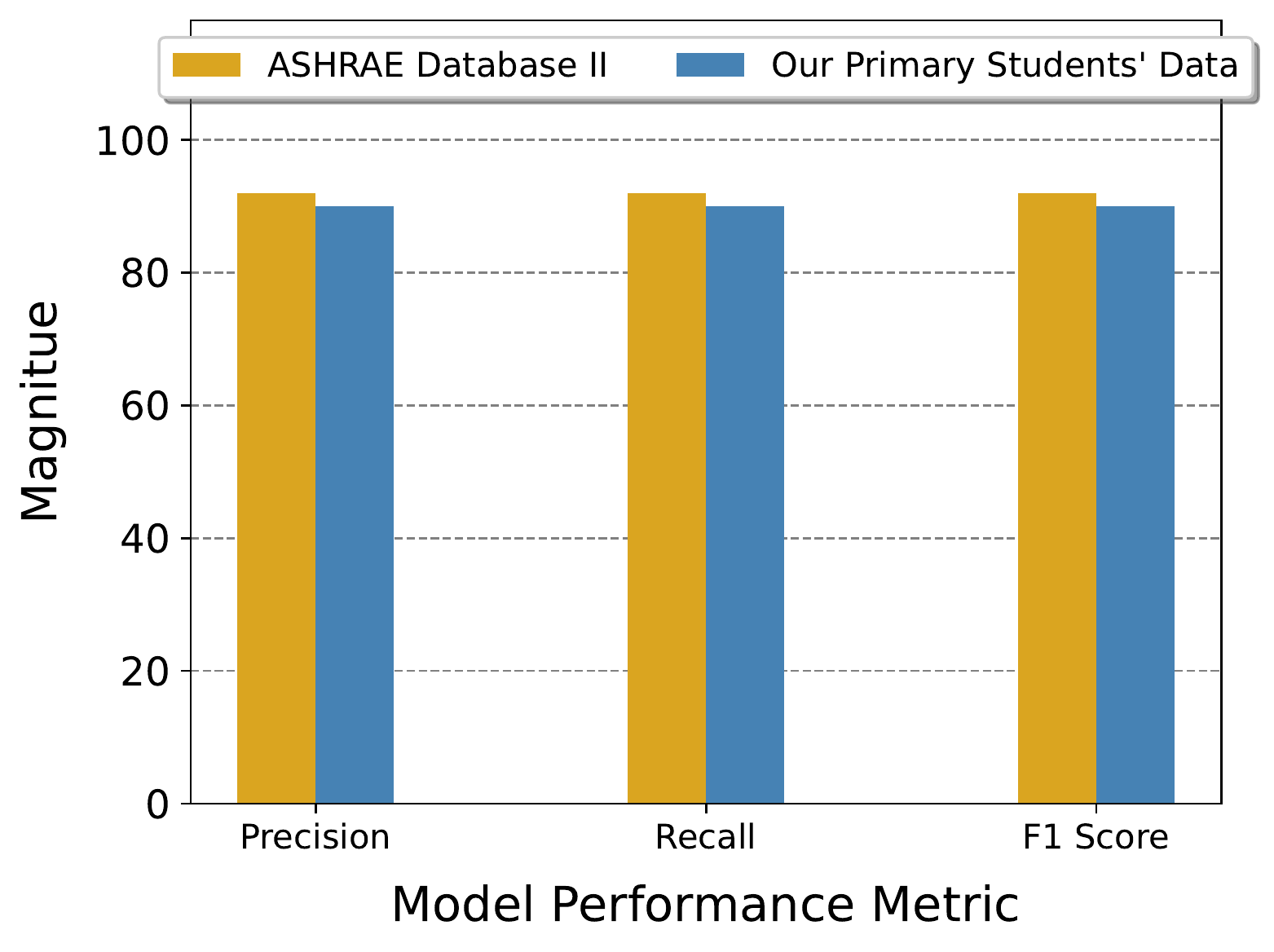}}\hfill%
	\subfloat[TPV] {\includegraphics[width=.33\linewidth]{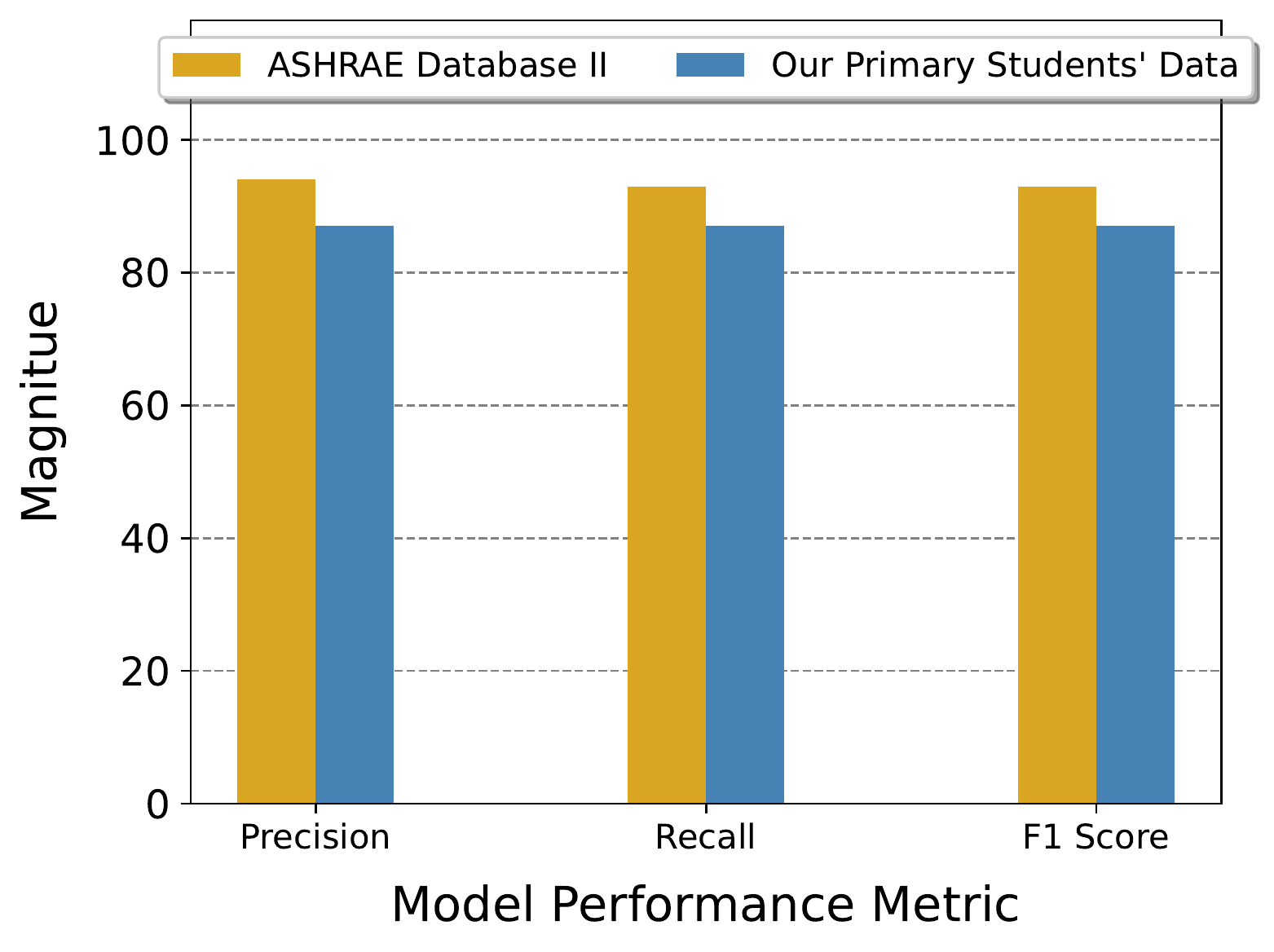}}\hfill
	\subfloat[TCV] {\includegraphics[width=.33\linewidth]{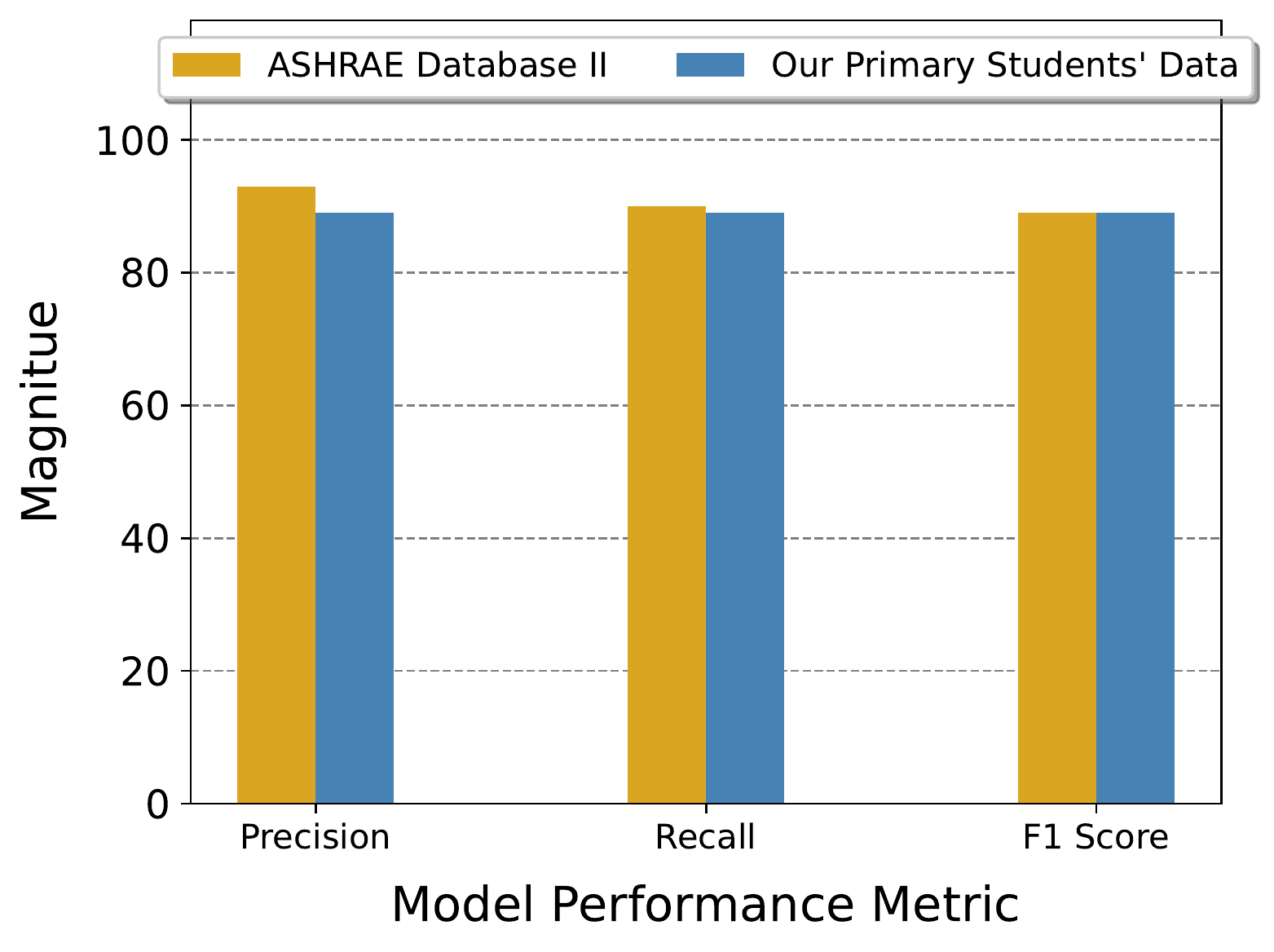}}\hfill%
\end{tabular}
    %\vspace*{0.1cm}
  \caption{\sys{} Performance Evaluation on Our Primary Students' Data and ASHRAE II Database }  
    \label{fig:DeepEval}
    %\vspace*{-0.4cm}
\end{figure*}
\subsection{Model Design and Implementation} \label{sec:model_creation}
% In this section, we present the details of the different modules of \sys{}.
 
 We define the prediction of TPV value as a task that require learning and similarly for TSV and TCV. Thus, the aimed prediction model is expected to maximize the likelihood of correct prediction of the joint performance of all targeted tasks at the same time given the input features. This is different from the state-of-the-art techniques that build a single model for each task and thus yields confusing predictions (e.g., too cold and very comfortable).
 
\sys{} employs multitask learning of different thermal comfort metrics. The intuition behind this is to effectively and simultaneously boost the learning ability of the trained model  for all target thermal comfort metrics leveraging the inherent correlation between them. This leads to a general model that jointly improves the prediction accuracy of each individual metric (task) as well as avoids the model overfitting while training. The state-of-art techniques, usually,  build a single model for each target thermal comfort metric which in general leads to overfitting problem and/or partial estimation of thermal comfort.

%  A deep fully-connected neural network is adopted here due to its representational ability, which allows learning of complex patterns

\sys{} adopts hard parameter sharing which is the most commonly used approach to training multitask neural networks \cite{ruder2017overview}. It is generally applied by sharing the hidden layers between all tasks, while keeping several task-specific output layers. Hard parameter sharing greatly reduces the risk of overfitting. In fact,  it showed that the risk of overfitting the shared parameters is an order $T$ -- where $T$ is the number of tasks -- smaller than overfitting the task-specific parameters, i.e. the output layers. This makes sense intuitively: The greater the number of tasks that are learned simultaneously, the more the proposed model has to find a representation that captures all of the tasks, thereby reducing the chance of overfitting. % in the original task.

To formally state the of proposed multi-task learning model, assume there are $T$ tasks. For each task $t$, we have $N$ samples;  $s^t_i= (x_i,y^t_i)$ denotes to the $i^{th}$ sample, where $i\in N$, $x_i$ is the set of features and $y^t_i$ is corresponding label of the $t^{th}$ task. A task is an abstraction read from raw data.
Typically, each task $t$ has a set of training samples which overlaps with the other tasks in the input features. The most traditional way is to train an individual model $f_\theta^t(x,y^t)$ for each task $t$ that maximizes the probability $P(y^t|x)$ of obtaining the true label $y^t$ given the input $x$, where $\theta$ is the model parameters. However, we build multitask learning over all tasks leading to a tasks-collaborative prediction model $f_\theta(x,y)$ where $y = \{y^1,y^2,.., y^T\}$. This ensures that obtained model is  more robust since it optimizes the cumulative prediction performance of all tasks together.

Figure~\ref{fig:network} shows the proposed deep network structure. We construct a deep fully connected neural network consisting of a common cascaded hidden layers of non-linear processing neurons. 
Specifically, we use the hyperbolic tangent function (tanh) as the activation function for the hidden layers due to its non-linearity, differentiability (i.e. having stronger gradients and avoiding bias in the gradients), and consideration of negative and positive inputs~\cite{lecun2012efficient}. 
The input layer of the network is also common with a vector of length $d$ representing the collected features from the students in the school of interest.
\color{black}
 The network consists of three subnetworks  stacked over the common hidden layers; each subnetwork is dedicated to one of the thermal comfort metrics (TSV, TPV and TCV). 
 The number of neurons at the output of each subnetwork is corresponding to the number of comfort levels (values) of its dedicated metric. For instance, thermal comfort can be reported by three levels of TSV including, -1,0,1 for cool, neutral, and warm.  Thus the TSV subnetwork is trained to operate as a  multinomial  (multi-class) classifier by leveraging a softmax activation function in the output layer.  This leads to a probability distribution over the reference TSV levels given an input. The same architecture is considered for TPV and TCV.
 
To increase the model resilience to over-fitting, \sys{} employs the \textbf{dropout regularization} ~\cite{srivastava2014dropout}  which is shown to be feasible for the efficient training of deep neural networks. This technique can sample from many neural networks of different architectures during the training process. This can be realized by stochastically removing (i.e. dropping out) some neurons in addition to their connections from each layer in the network. 
In effect, each layer has a new “view” different from the original configured layer in each epoch in the training.
Dropout has the effect of making the training process noisy, forcing units within every layer to stochastically take on more or less responsibility for the inputs.  As a result, it prevents the neurons from co-relying on each other during the training process, leading to a more robust model that is less likely to overfit the training data.
% We provide an experiment to show the effect of dropout on the results in Section~\ref{sec:dropout_results}.

The current implementation and validation of \sys{} is done locally, and both training and testing of the models ability to predict students' thermal comfort perception is performed in the offline mode. The proposed deep learning model was implemented using 
%training using Python programming language and Keras learning library. 
\textit{Keras}, which is a high-level neural networks API running on top of the Google TensorFlow framework~\cite{tensorflow}. The model is trained on a Lenovo ThinkStation P920 server with Nvidia RTX3080 ti GPU, and 320GB RAM.
After running numerous experiments, the deep learning architecture of 20$\times$50$\times$80$\times$100$\times$120, delivered the best performance.

% \begin{table}
% \centering
% \caption{Notations used in the paper.}
% \label{table:notations}
% {\renewcommand{\arraystretch}{1.3} 
% \begin{tabular}{p{1.5cm}p{7cm}}
% \hline
% \textbf{Notation} &\textbf{Description} \\ \hline 
%  $N$ & Number of samples. \\    
%  $T$ & Number of tasks (i.e, thermal comfort metrics). \\    
%  $y^t$ & The label of the $t^{th}$ task. \\    
%  $d$& The number of input features.\\
% \hline        
% \end{tabular}
% }
% \end{table}

\begin{figure*}[htbp]
 \centering%
\begin{tabular}{cc}
    \subfloat[Impact of Layers] {\includegraphics[width=.33\linewidth]{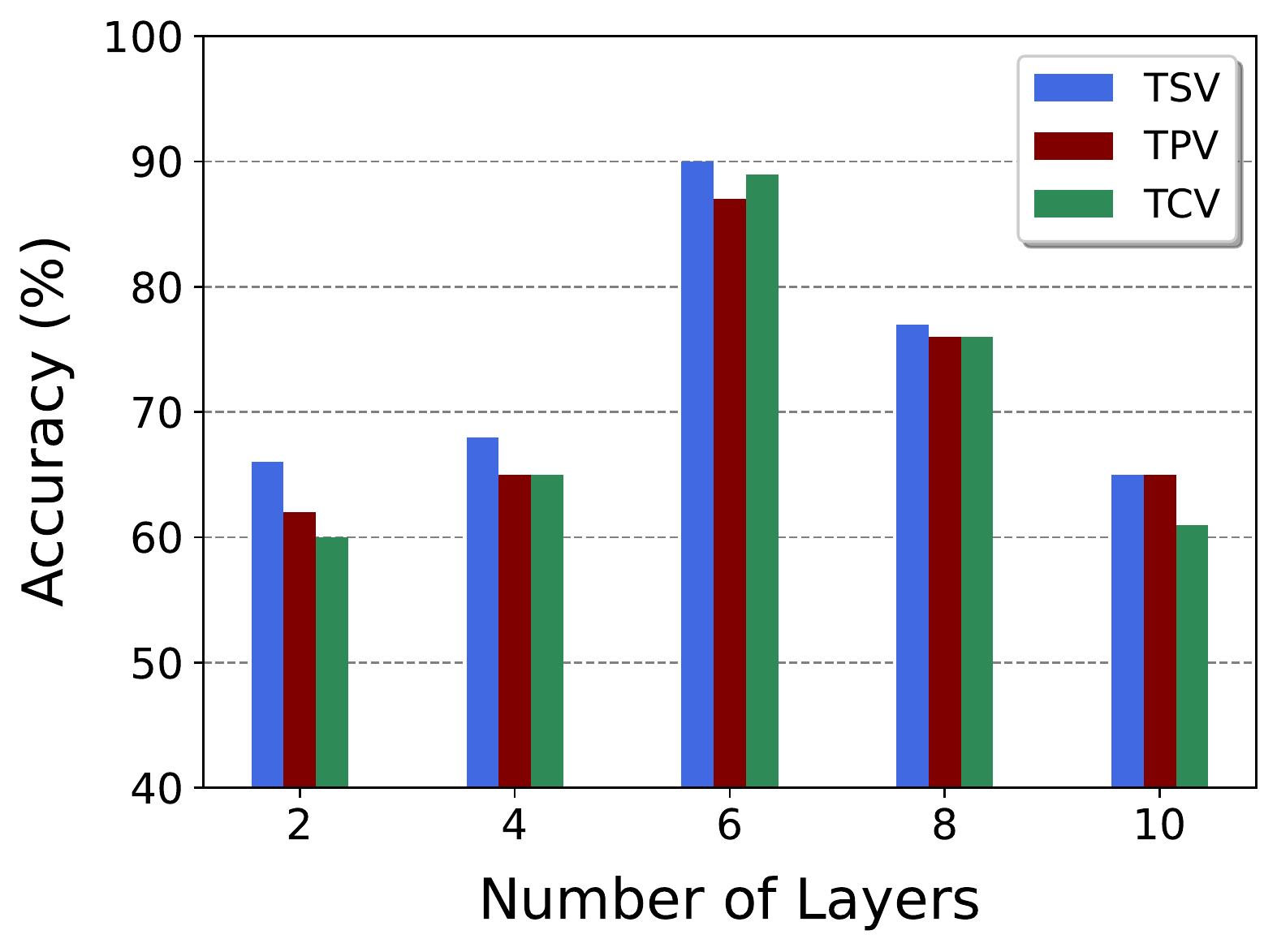}
    \label{fig:eval_layers}
    }\hfill%
	\subfloat[Impact of Epochs] {\includegraphics[width=.33\linewidth]{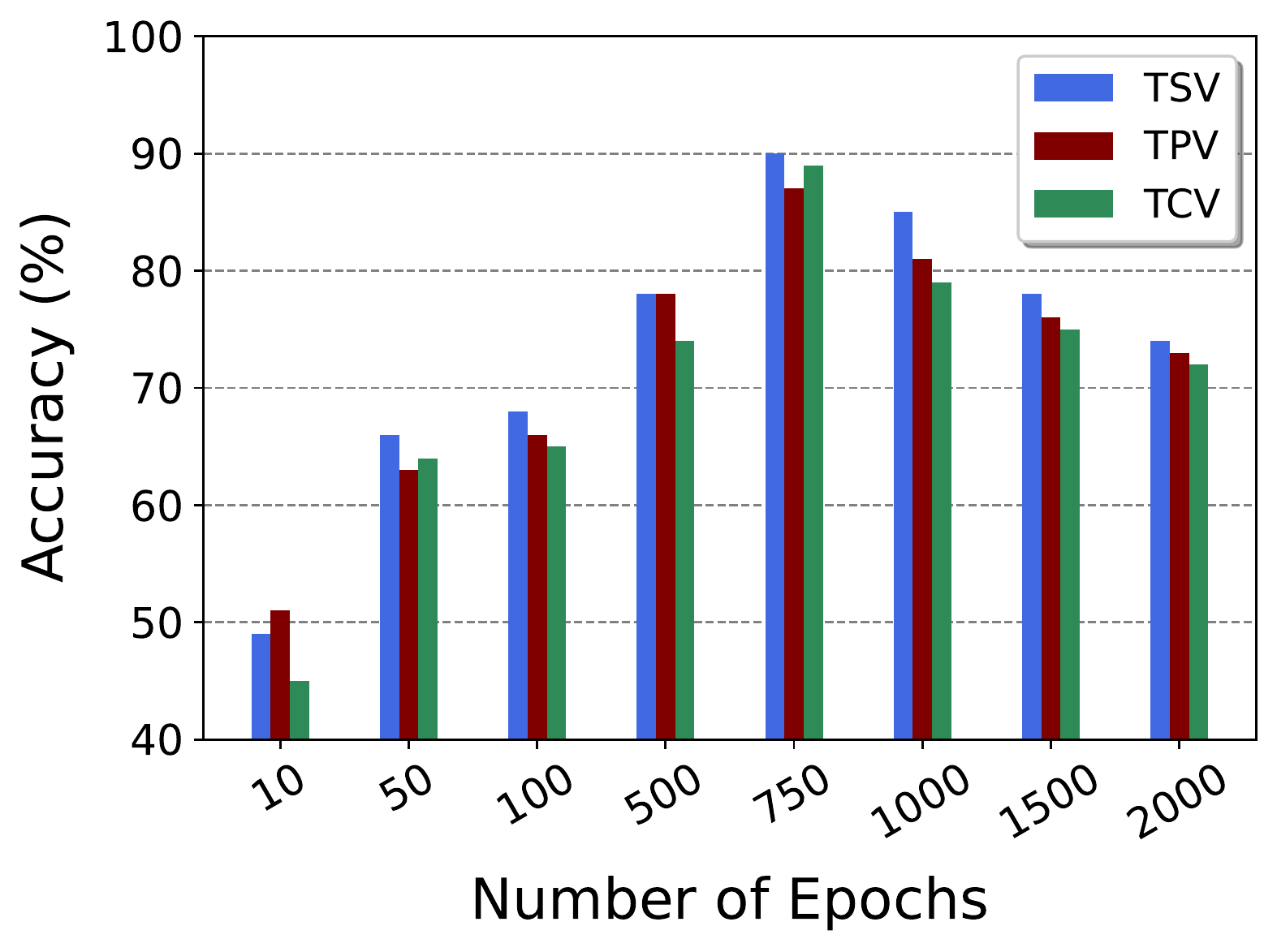}
	\label{fig:eval_epocs}
	}\hfill
	  \subfloat[Impact of Learning Rate] {\includegraphics[width=.33\linewidth]{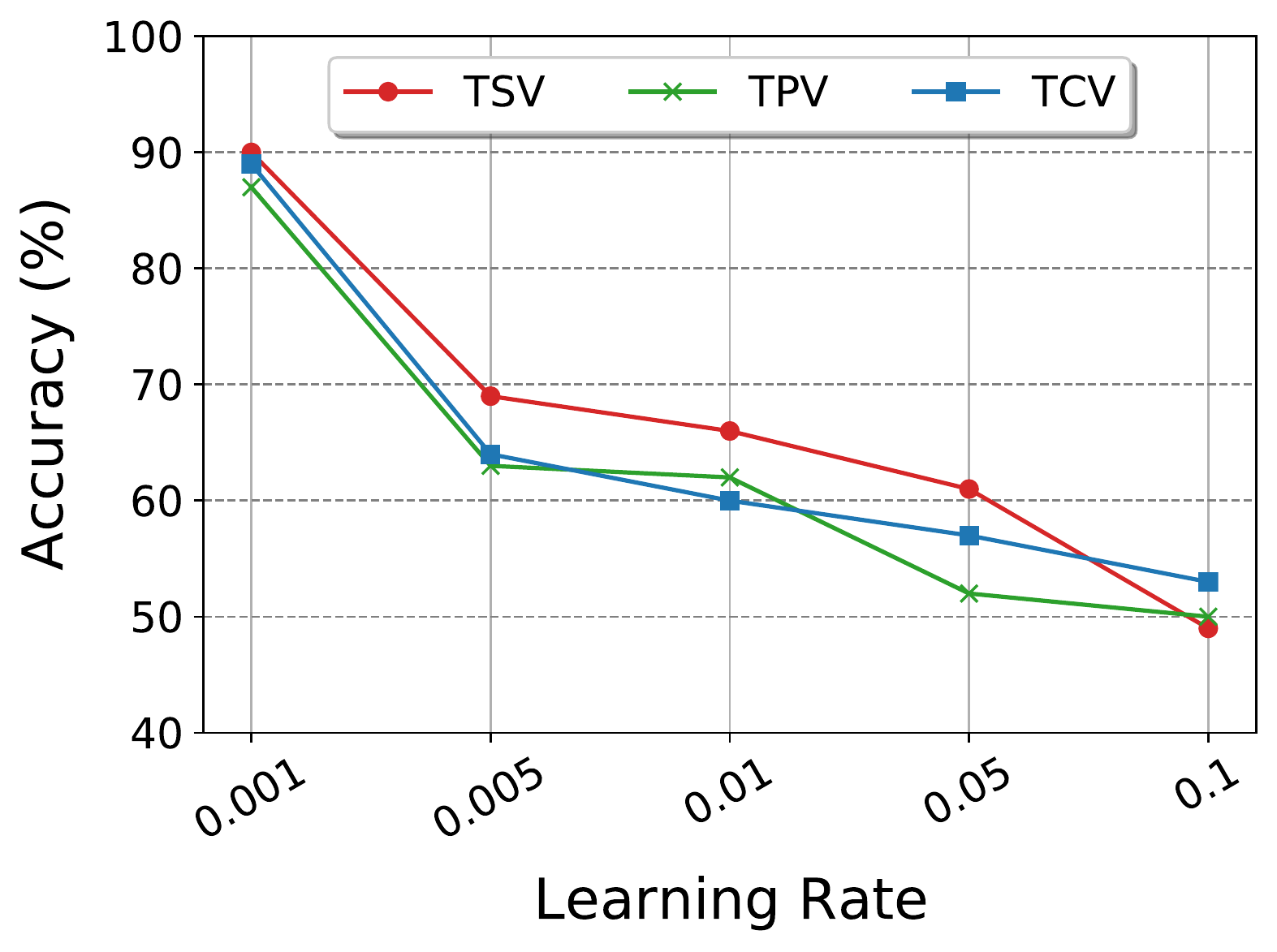}\label{fig:eval_learing_rate} 
	  }\hfill%
\end{tabular}
    %\vspace*{0.1cm}
  \caption{Effect of Varying Hyperparameters on \sys{} Performance} 
    \label{fig:hyper}
    %\vspace*{-0.4cm}
\end{figure*}

\section{Evaluation of DeepComfort }  \label{sec:evaluation}
This section presents the performance evaluation of the proposed \sys{} multi-task model and highlights the challenges involved in thermal comfort prediction for primary school students. 
\subsection{Evaluation Methodology}
To evaluate the proposed system and its trained multi-task learning model and confirming its generalization ability, K-fold cross-validation is employed, where $k=5$. The dataset is partitioned into $k$ subsets i.e., folds. Each time, $k-1$ folds are merged to form a training set and the remaining one is leveraged as the validation set. Hence, every sample of our dataset appears in a validation set exactly once and appears in a training set $k-1$ times. Thereafter, the average error across all $k$ folds is reported and is used to select the model parameters. This significantly reduces the impact of the bias-variance problem due to the interchange of the training and validation sets. 

In this section we quantify \sys{}'s performance using different criteria including, Accuracy, Precision, Recall, F-Measure, and Confusion Matrix.
Accuracy is the percentage of predictions our model correctly obtained.
% given as: 
% %&
% %\begin{equation}
% $
% \text { Accuracy }=  
% % \frac{Number of correct predictions}
% % {Total number of predictions } \\ =  
% \frac{t p+t n}{t p+t n+f p+f n}
% $.
%\end{equation}
Precision quantifies the number of correct instances out of all  predictions as an arbitrary class.
% that actually belong to the positive class
% , as: $\text { Precision }=\frac{t p}{t p+f p}
% $.
Recall quantifies the number of correct  predictions made out of all instances of a specific class.
% , as: $\text { Recall }=\frac{t p}{t p+f n}$.
F-Measure (F1-score) provides a single score that balances both the concerns of precision and recall in one number, as: $\mathrm{F} 1 =2 \times \frac{\text { Precision } * \text { Recall }}{\text { Precision }+\text { Recall }}$.
% Finally, 

 The overall validation and performance evaluation of \sys{} is done using Precision, Recall, F-Measure, and Confusion Matrix. However, for the clarity of presentation, the effect of different system parameters and categorical features is presented in terms of Accuracy. 

%\color{red}
%{}
\begin{table*}[b]
    \centering
   \small
    \caption{DeepComfort vs.Single-task Learning Algorithms}
    \label{tab:MTLSTL}
    %\begin{tabular}{@{}nd{1.1}*{3}{d{1.2}}d{1.1}d{3.2}@{}}
    %\begin{tabular}{CCCCCCCCCC}
    \begin{tabular} {cccccccccc}
      \toprule
      \multirow{4}{*}{Machine Learning Techniques} &
        %\multicolumn{1}{V{12em}}{Machine Learning Techniques} &
        % \multicolumn{1}{}{} &
        \multicolumn{9}{c}{Thermal Comfort Output Metrics} \\
        \cmidrule(lr){2-10}
        &
        %\multicolumn{1}{N@{}}{Preis} \\
        %&
       % \multicolumn{5}{N}{Leiterspannung an der Einbaustelle} \\
      %\cmidrule(lr){2-10}\\
        \multicolumn{3}{v{15em}}{Thermal Sensation Vote} &
        \multicolumn{3}{v{15em}}{Thermal Preference Vote} &
        \multicolumn{3}{v{15em}}{Thermal Comfort Vote} \\
        %\multicolumn{3}{V{6.5em}}{Thermal Comfort Vote} &
        %\multicolumn{1}{V{4em}}{Nenn"-spannung} \\
        \cmidrule(lr){2-4}\cmidrule(lr){5-7}\cmidrule(lr){8-10}
        &
    
        \multicolumn{1}{v{4.5em}}{Precision} &
        \multicolumn{1}{v{4.5em}}{Recall} &
        \multicolumn{1}{v{4.5em}}{F1-score} &
        \multicolumn{1}{v{4.5em}}{Precision} &
        \multicolumn{1}{v{4.5em}}{Recall} &
         \multicolumn{1}{v{4.5em}}{F1-score}&
         \multicolumn{1}{v{4.5em}}{Precision} &
        \multicolumn{1}{v{4.5em}}{Recall} &
         \multicolumn{1}{v{4.5em}}{F1-score}\\
        % &
        % \multicolumn{1}{N}{\unit{kV}} &
        % \multicolumn{1}{N}{\unit{kV}} &
        % \multicolumn{1}{N}{\unit{kV}} &
        % \multicolumn{1}{N}{\unit{kV}} &
        % \multicolumn{1}{N}{\unit{kV}} &
        % \multicolumn{1}{N}{DM} \\
      \cmidrule(l){1-1}\cmidrule(l){2-2}\cmidrule(l){3-3}\cmidrule(l){4-4}%
        \cmidrule(l){5-5}\cmidrule(l){6-6}\cmidrule(l){7-7}\cmidrule(l){8-8}\cmidrule(l){9-9}\cmidrule(l){10-10}
SVM	&	67	&	66	&	55	&	43	&	48	&	44	&	50	&	60	&	51	\\
Random Forest	&	55	&	58	&	56	&	45	&	46	&	46	&	53	&	57	&	55	\\
Decision Tree	&	58	&	55	&	56	&	48	&	48	&	48	&	51	&	52	&	52	\\
KNN	&	53	&	53	&	53	&	44	&	45	&	44	&	51	&	51	&	51	\\
AdaBoost	&	58	&	62	&	54	&	39	&	38	&	37	&	34	&	59	&	43	\\
DNN	&	67	&	67	&	67	&	64	&	64	&	63	&	63	&	64	&	64	\\
\textbf{DeepComfort}	&	90	&	90	&	90	&	87	&	87	&	87	&	89	&	89	&	89	\\
      \bottomrule
    \end{tabular}
  \end{table*}
  
\begin{figure*}[h]
\centering 
    \includegraphics[width=\linewidth]{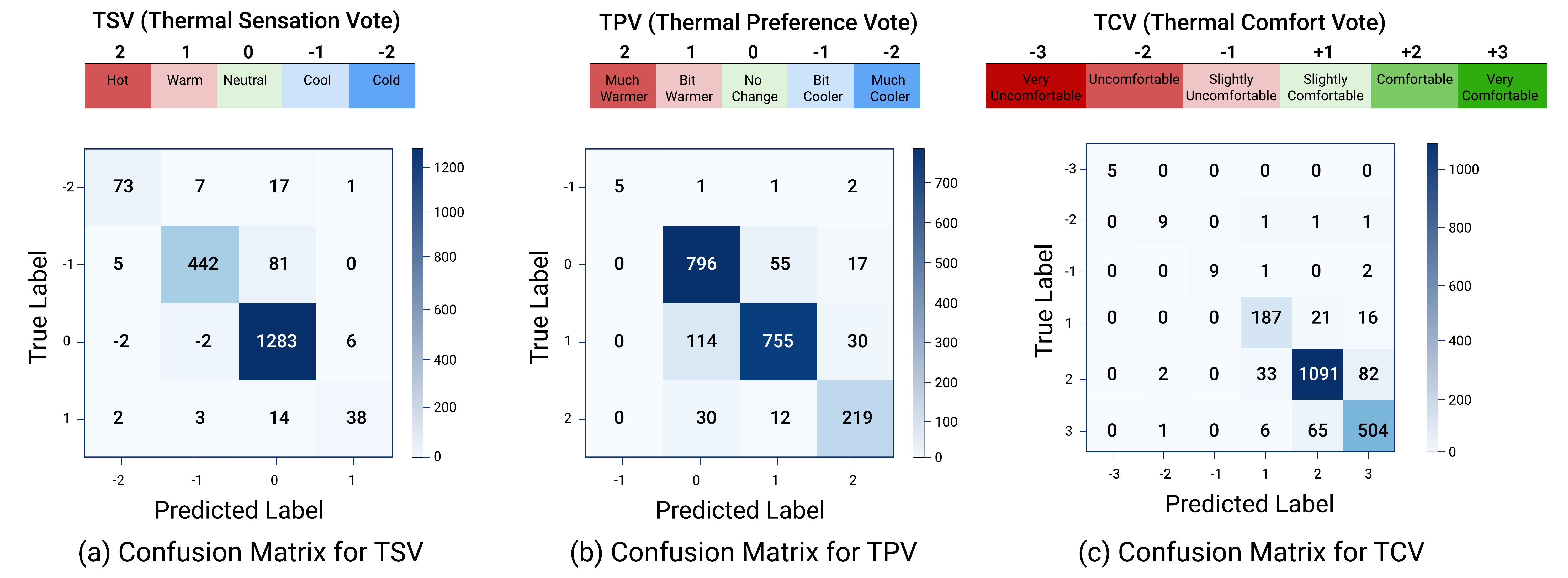}
\caption{Confusion Matrix for Individual Classes in the Multi-Task Model}
\label{fig:confusion_matrix}
\end{figure*}
\subsection{Validation on ASHRAE II \& Our Data}\label{sec:ashrae} 
\manas{The first step is to evaluate the generalization ability of the proposed \sys{} model. To that end, the model is trained and tested on the ASHRAE II dataset \cite{db2}. The ASHRAE Global Thermal Comfort Database II is the largest publicly available open-source database, created from landmark thermal comfort field studies in 28 countries, spread across the globe. The database includes over 50 attributes, including objective environment data, subjective TC metrics, built environment characteristics, climate and weather data, and participant information \cite{db2}. However, the ASHRAE II database doesn't have any dataset for primary school students, i.e., for students of age 14 or lower. Therefore, all data for students under the age of 18 in naturally ventilated classrooms available in ASHRAE II database available was considered for evaluation of \sys{}.}

Figure~\ref{fig:DeepEval} shows the performance of \sys{} benchmarked against ASHRAE II data. There are some characteristic differences between the two datasets, which include the number of features, the number of classes of each TC metric, and the number of samples. The comparison is performed in terms of Precision (Accuracy), Recall, and F-score metrics. 
Despite the differences in the two test-sets, the \sys{} system demonstrates a consistent performance for all three metrics. The, prediction performance is slightly better for ASHRAE II database. This is justified, as ASHRAE II data participants are young adults (ages 14 to 18) or adults with developed cognition and reasoning. In contrast, our primary student data is mainly gathered from participants in the age range of 6 to 13, and due to children's cognitive limitations, is likely to have a higher frequency of ``illogical votes,'' which can be considered to be outliers but can not be ignored or dropped from the model. 
Nevertheless, \sys{} overcomes this challenge and demonstrates high prediction Accuracy for all three TC response metrics.  The results validate the suitability of multitask learning for thermal comfort prediction even when outliers are present.  
%\begin{slide*}
  
%\end{slide*}

\subsection{Impact of Hyperparameters} \label{sec:parameters}
Hyperparameter tuning is vital for a deep neural network's performance. Recent surveys on application of machine learning to the domain of thermal comfort have expressed a concern that a deep neural network may become a \textit{black box} for the research community, if the inner workings of the models are unknown \cite{ML_TC_REVIEW_1, ML_TC_REVIEW_2}. It also poses challenges in replicating the proposed models. To address these concerns, this section analyzes the impact of different hyperparameters on \sys{} performance viz., the number of layers in the network, the number of epochs, and the learning rate.

\subsubsection{Number of Layers}
One of the salient hyperparameters is the number of layers of the deep network as it reflects the distributed (i.e. hierarchical) capability of the model.
Figure~\ref{fig:eval_layers} shows the effect of changing the number of hidden layers of the multitask model on \sys{} accuracy.  The figure shows that increasing  the number of layers increases the performance due to increased model capacity until it reaches six layers. After that, the model tends to overfit the training data, reducing performance in all TC metrics. Therefore, we choose six layers as the default number of layers in our multitask thermal comfort model. 

\subsubsection{Number of Epochs}
Training a deep network is a challenging process since overtraining may force the model to stop generalizing and  learn to memorize the training data. On the other hand, too little training may lead to underfitting leading to  poor performance even on the training set. Figure~\ref{fig:eval_epocs} shows that 750 epochs is an optimal value which leads to the best performance of \sys{}. 

\subsubsection{Learning Rate ($\alpha$)}
Tuning the model's learning rate is an important step as it controls how much the network weights are adjusted with respect to the loss gradient.
Figure~\ref{fig:eval_learing_rate} shows the impact of changing the learning rate on the \sys{} performance. 
The figure shows that a learning rate of $\alpha$ = 0.001 obtains the best performance of \sys{} in all thermal comfort metrics. 
This can be justified as the model at this value balances between larger and smaller learning rates.
Larger learning rates may lead to a divergent training process. On the other hand, smaller learning rates may conversely lead to non-optimal convergence of the training process.

\subsection{DeepComfort vs. Single-task Models}
%\color{blue}
An important aspect of evaluation is to compare the performance of \sys{} with single-task thermal comfort prediction models implemented using state-of-the-art ML techniques. A total of 6 single-task models are considered which include both supervised and unsupervised, shallow and deep algorithms, briefly described below.

In \cite{wang2020b} Support Vector Machine (SVM) is employed to implement single-task models that predict the TSV and TCV individually. Further, a deep neural network (Bayesian  Network) approach, denoted as "DNN", is adopted in \cite{cakir2022bayesian} for estimating TSV. \sys{} is also compared with Decision Tree which is commonly used technique for single-task TC classification and prediction \cite{liu2019a}. Additionally, Random Forest classifier is also included in the comparison due to its effectiveness as it builds a forest of many decision trees; each of them outputs a class prediction, and the class with the majority votes will be reported by the model \cite{liu2019a}. In K-nearest neighbor (KNN), a class is estimated by its plurality among its neighbors, i.e., the sample is assigned to the class most common among its ``k'' nearest neighbors \cite{shan2020b}. Finally, the Adaptive Boosting (AdaBoost) technique is an ensemble boosting classifier. It builds a robust classifier by combining multiple weak classifiers ensuring accurate predictions of unusual samples \cite{ML_TC_REVIEW_2}. 

Table~\ref{tab:MTLSTL} shows \sys{}'s performance when compared to these single-task techniques with respect to F1-score, Precision, and Recall. F1-score is chosen for comparison instead of Accuracy, because due to the data imbalance in the dataset, prediction of minority classes is not adequately reflected in Accuracy. The results confirm that the proposed multi-task learning model outperforms the other techniques, even the Bayesian deep neural network. 
This can be justified for to two primary reasons. First, the distributed learning ability of the deep neural network enables automatic feature learning improving the accuracy in prediction. Second, the enhanced generalization ability due to the multi-task learning model further maximizes the prediction accuracy of all tasks simultaneously. The second reason is the main point of distinction when compared to the 6 single-task models. Moreover, the proposed MTL model's design boosts the learning of general and accurate models through regularization techniques.
\color{black}

\begin{figure*}[htbp]
 \centering %
\begin{tabular}{cc}
   % \subfloat[ Feature Importance in \sys{} Performance ] {\includegraphics[width=.45\linewidth]{Plots/Features.pdf}}\hfill
    \subfloat[ School Architecture \& Classroom Environment ] {\includegraphics[width=.36\linewidth]{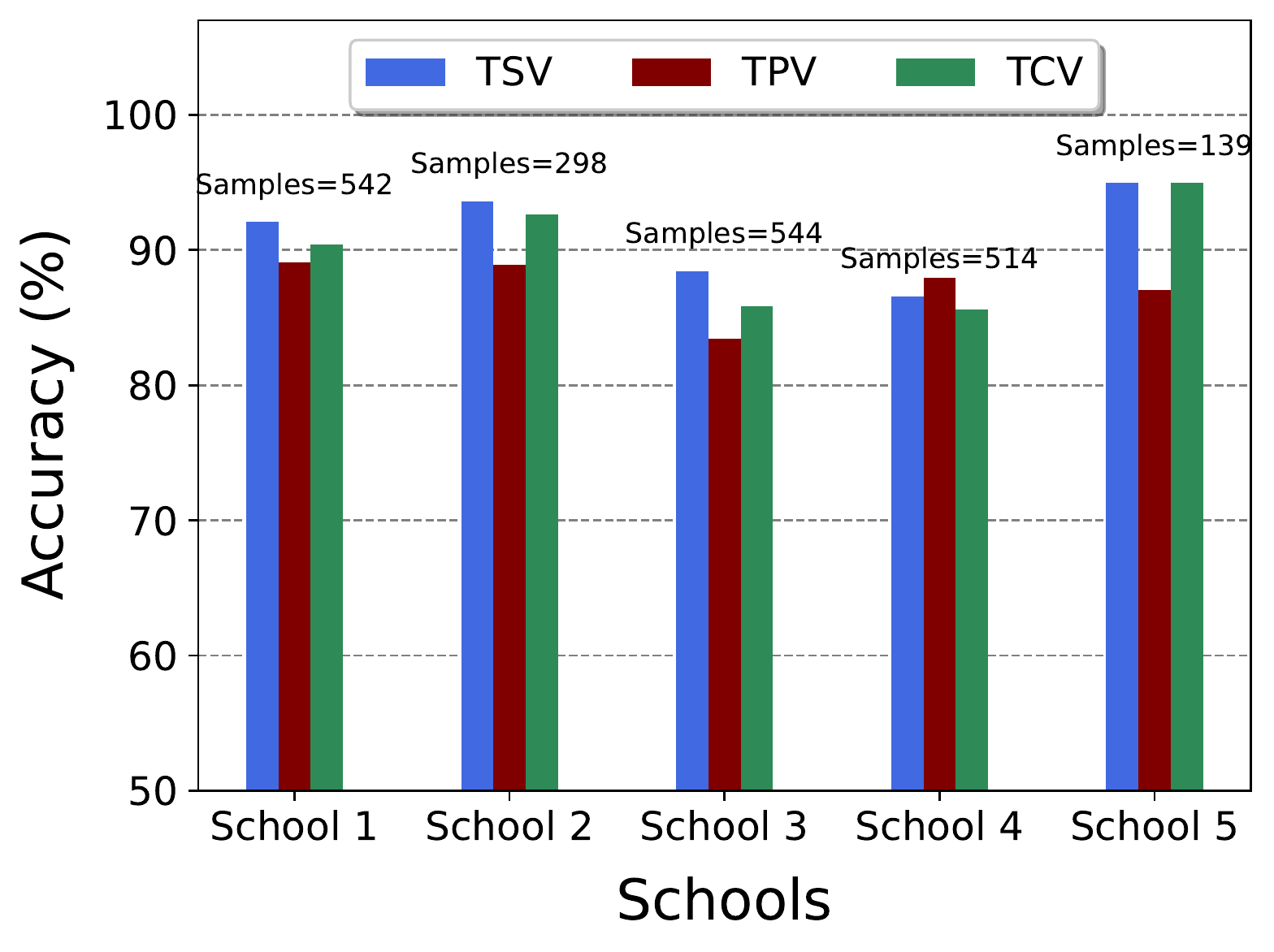}}\hfill
      \subfloat[ Gender \& Sociophysiological Aspects ] {\includegraphics[width=.28\linewidth]{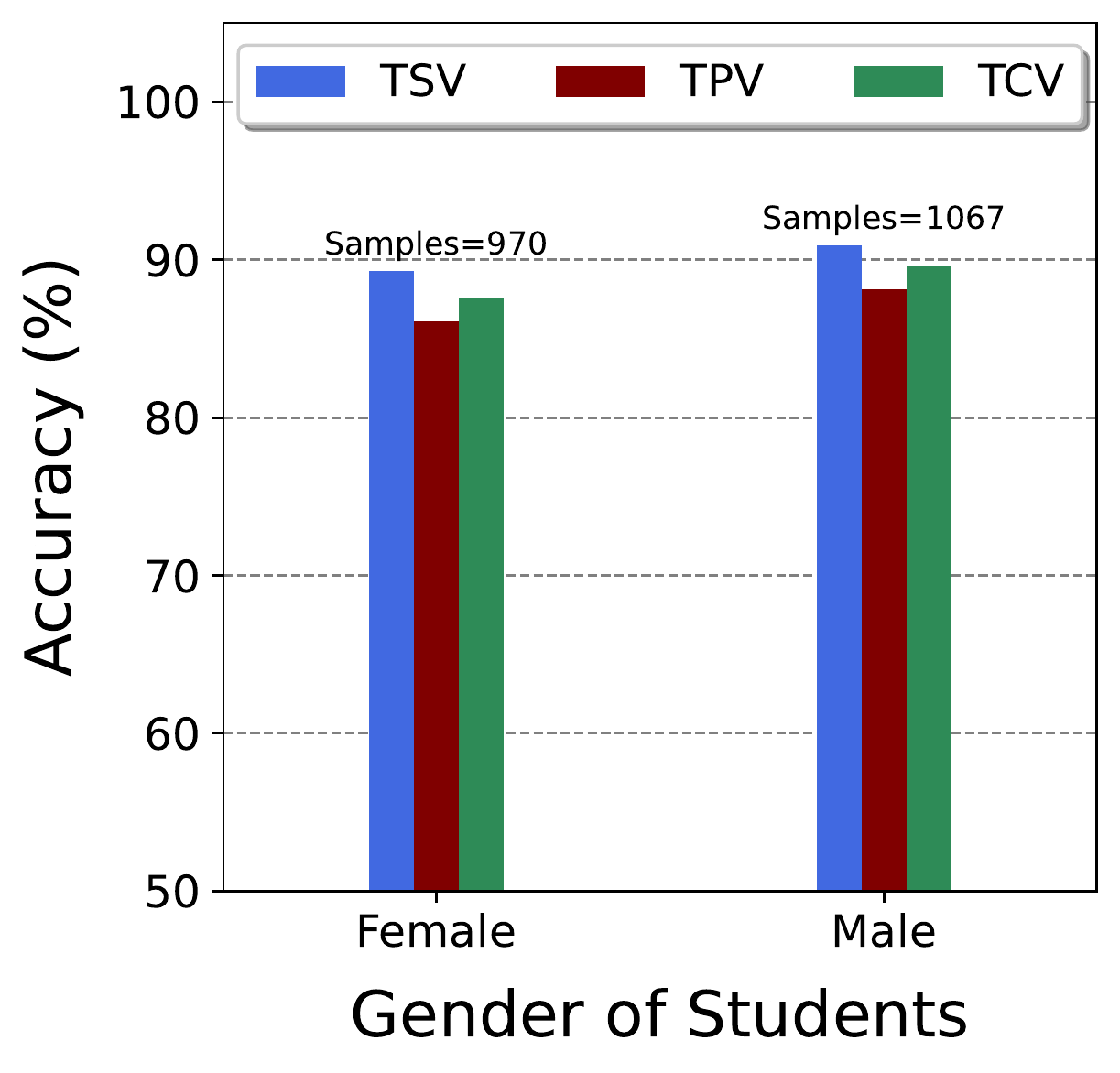}}\hfill% 
    \subfloat[Grade \& Cognitive Abilities ] 
    {\includegraphics[width=.36\linewidth]{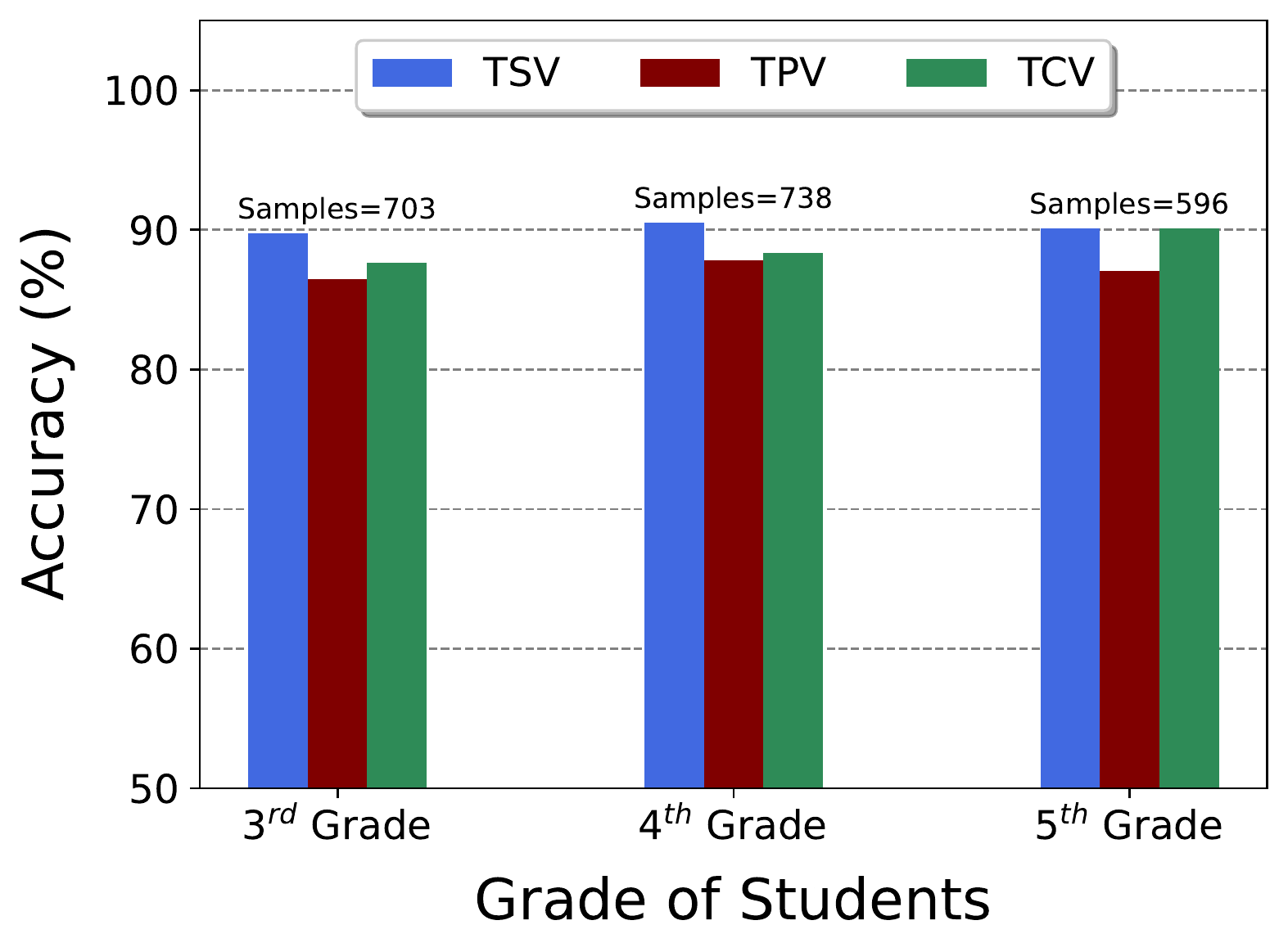}}\\
     \subfloat[ Survey Duration \& Attention Deficit ] {\includegraphics[width=.4\linewidth]{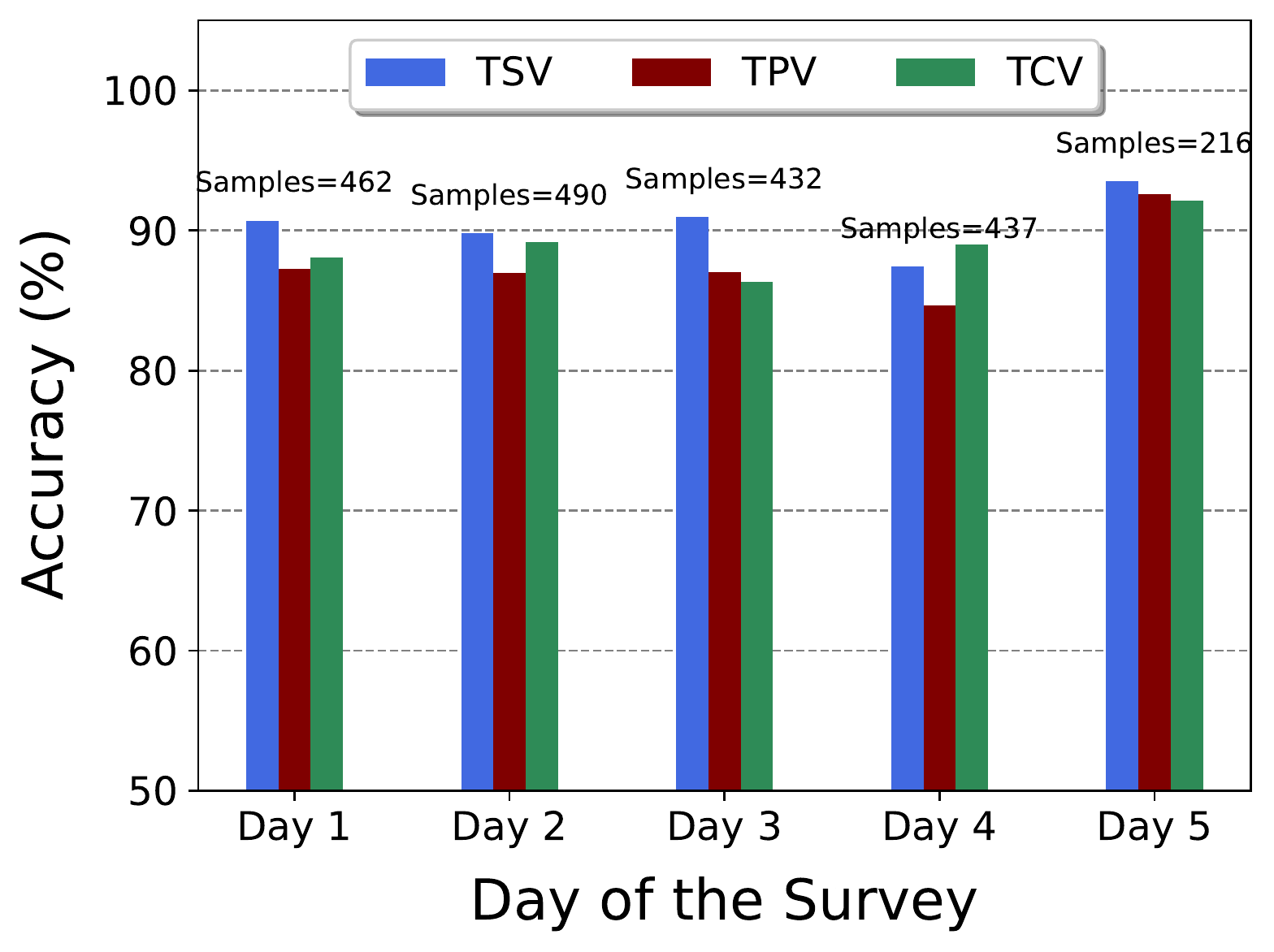}}%
	\subfloat[Survey Timings \& Weather] {\includegraphics[width=.4\linewidth]{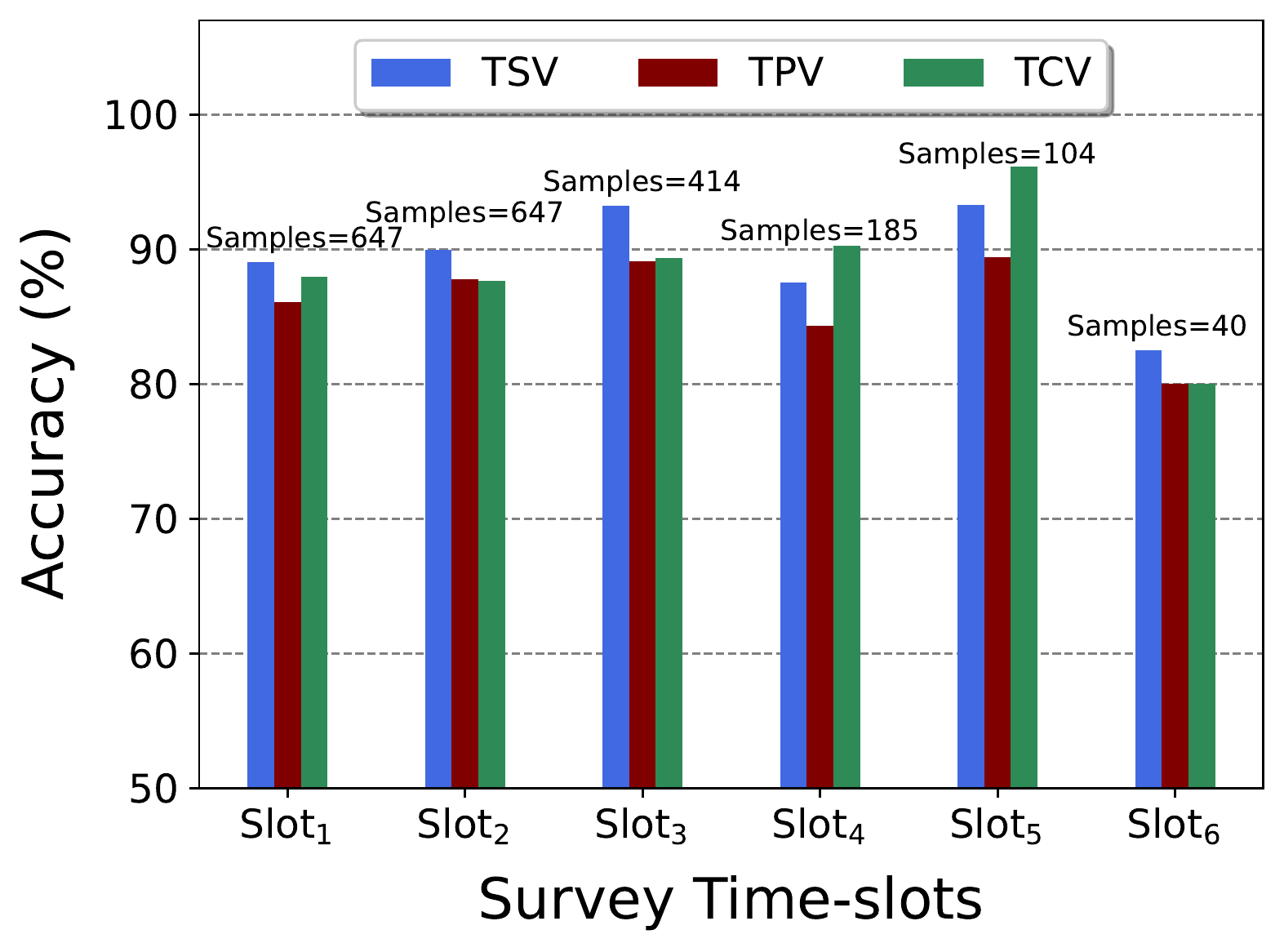}}\hfill\\
\end{tabular}
    %\vspace*{0.1cm}
  \caption{Impact of Features on the Consistency in \sys{} Performance } 
    \label{fig:consistency}
    \vspace*{-0.4cm}
\end{figure*}
\begin{figure}[h]
\centering 
    \includegraphics[width=0.9\linewidth]{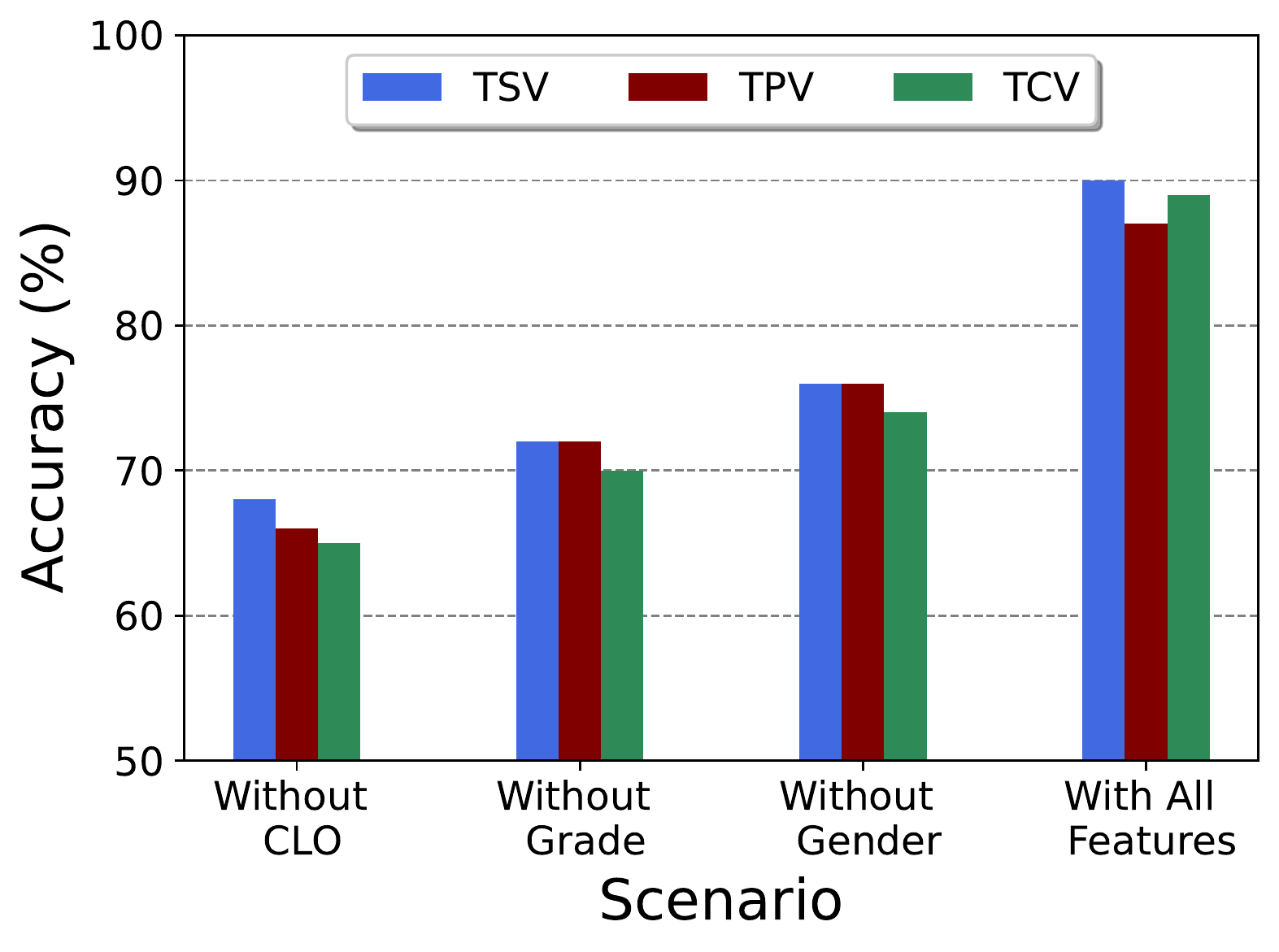}
    \caption{Impact of Features on \sys{}}
\label{fig:features}
\end{figure}

\subsection{Classification Challenges \& performance}~\label{sec:performance}
Next, the analysis of multi-class classification performance of \sys{} is presented. Figure~\ref{fig:confusion_matrix} shows the confusion matrices of the system performance corresponding to each \textit{task} (e.g., TSV, TPV, and TCV). Relevant sections of each TC metric scale are also presented for ease of reference.

The confusion matrices show the robustness of the system in dealing with data bias. For example, since the proposed model is trained with winter data in the composite climate, the majority of samples for TSV are either ``Neutral'' or ``Cool''. This biased data distribution, in general, leads to a biased model that incorrectly estimates other values such as ``Cold'' or ``Warm'' as ``Neutral'' or ``Cool''. Despite this challenge, the model shows a good generalization and non-biased capability by leveraging the generalization ability of multi-task learning and through regularization techniques. A similar non-biased performance, across all classes, can be observed for the other two metrics, viz., TPV and TCV in Figures~\ref{fig:confusion_matrix}~(b)~\&~(c).

It is noteworthy that some false predictions emerge from the unclear boundaries between different classes, e.g., ``Slightly Comfortable'', ``Comfortable'', and ``Very Comfortable'' in TCV. The role of cognition is crucial here, which highlights why TC prediction for primary school students in naturally ventilated environments is challenging. Despite the use of illustrations and easy-to-understand language, primary school students seem to find difficulty in assessing their situation with respect to the standard TC metric scales (TCV in this case). Multiple classes with nuanced differentiation achieved through the use of qualifiers such ``Slightly'' or ``Very'' seems to confuse young primary students. This lack of clarity also affects the correspondence (if not correlation) between student responses that fall in minority classes for the three metrics. Case in point, in Figure~\ref{fig:confusion_matrix}, the number of ``Cold'' sensation (TSV=-2) labels, are far less than ``Much Warmer'' preference (TPV=2) labels, both of which are much higher than ``Uncomfortable'' and ``Very Uncomfortable'' comfort labels (TCV=-2,-3). Thus, the inconsistency in responses while filling the survey (the model ground-truth), often results in inaccurate model predictions.
Yet, despite these challenges in the multi-task multi-class classification goal, the \sys{} system is able to simultaneously achieve a prediction Accuracy of  90\%, 87\%, and 89\%, for TSV, TPV, and TCV metrics, respectively.

\subsection{Categorical Features \& Model Accuracy}
%In Section~IV, a detailed analysis of various factors or features was presented. 

Figure~\ref{fig:features}, demonstrates the impact of various features such as clothing, grade, and gender on \sys{} performance. Two observations can be made. First, the features invariably impact the performance of the a classification models. Second, the impact of the feature on a model's performance i.e., \textit{feature importance}, varies across features. 

It is a worthwhile objective for a classification/prediction model to have high generalization capability, i.e., achieve high accuracy regardless of the distribution of features. However, from the perspective of prediction of subjective TC responses, it is important to ensure that the model performance is consistent for all categories of a feature. 

Thus, it is desirable, and an objective of this work, to stabilize \sys{} performance across schools, survey timings, the gender of students, days of the survey, and the grade of students. \sys{} achieves this objective by precisely training the model for specific variations in the categorical features, e.g, Male and Female students. The complexity of the task lies in the trade-off involved between achieving generalization ability and maintaining high-accuracy for all values of categorical features.\footnote{An alternative goal can be to highlight the differences in the impact of individual features on students' TC perception, which is beyond the scope of this work.}

The results for the five features analyzed earlier are presented in Figure~\ref{fig:consistency}. \sys{} is able to achieve high Accuracy (80\%-96\%) in all three metrics, for all features in the feature-category-specific evaluation. There is some variation in Accuracy, which is expected. Further, explanations for the fluctuation in Accuracy in feature-categories can be attributed to the unusual distribution of TSV, TPV, and TCV values in those categories.

With respect to Schools as the feature, School~3 and School~4, have the lowest Accuracy. Considering the case of School~4, the fluctuation is due to the fact that a higher proportion of students respond to feeling ``Cold'' and ``Cool'' sensations (TSV=-2,-1), yet they also respond to feeling varying levels of comfort (TCV=1,2,3). This ambiguity is less prominent in other schools, where the majority of students experience a ``Neutral,''  sensation along with varying levels of comfort. In School~3, there exists a lack of congruence between the distribution of sensation, temperature preference, and comfort, votes. For example, School~3 has a high proportion of both ``Cold'' sensation votes (TSV=-2) and ``Very Comfortable'' (TCV=3) votes, which confuses the model, resulting in poor Accuracy. 

Considering female and male students as the categories for Gender, a higher proportion of female students responded that they feel ``Cold'' (TSV=-2) but did not express any discomfort in the TCV scale. As a consequence, the Accuracy for male students is slightly higher. A similar trend is observed in 3$^{rd}$, 4$^{th}$, and 5$^{th}$ grades as categories for the grade feature. As cognitive ability of the students increases with grade, the number of illogical votes go down, leading to slightly improved Accuracy. 

For days of the survey, the Accuracy results for specific days conform to the trend of TSV, TPV, and TCV distributions, discussed in Section~IV. The Accuracy for the first 3 days remains stable, with minor variations, but drops slightly on Day~4 and is highest for Day~5. Another reason for very low Accuracy is high data-imbalance. For example, for Slot~6 in survey time-slots, the Accuracy  drops down to 80\%, which is because of high data imbalance in this class. 

\par{}
The analysis presented in this section demonstrates that the proposed \sys{} model offers high generalization capability and stable performance. Based on the results and the challenges encountered, a few inferences and conclusions  are presented in the next section.

\section{Conclusions and Future Work}~\label{sec:conclusions}

This work sought to address the problem of multiple TC prediction models for each indoor space, one specific to each metric. To that end, it proposed a multi-task learning inspired deep learning model named ``DeepComfort'', that concurrently predicts three TC metrics, viz, TSV, TCV, TPV. Further, this work envisions a real-world implementation of the proposed DeepComfort MTL model. Thus, the model was validated on a large dataset gathered through a month-long comprehensive survey and field experiment involving 5 schools, 14 classrooms, and 512 unique primary school student participants.

The first inference can be made on the suitability of multi-task learning for thermal comfort prediction. The proposed MTL solution requires a single model to simultaneously predict the three subjective TC response metrics, viz., TSV, TPV, and TCV. Deepcomfort is shown to outperform 6 single-task learning models. Further, predicting thermal comfort for primary school students in naturally ventilated environments is challenging because of children's limited cognitive ability to perceive and assess the classroom environment. Consequently, there is a higher volume of illogical responses in the surveys that typically lower the accuracy of multi-class classification. Despite these challenges, the deep network architecture of \sys{} allows it to maintain high prediction Accuracy for our primary student data as well as ASHRAE II data, ensuring high generalization capability. The \sys{} model also demonstrates consistent performance for different categories of categorical-features with different characteristics. 

Given the satisfactory performance of the proposed multi-task learning model, the next is to extend the implementation to predict a larger set of TC metrics including Thermal Acceptability, Temperature Satisfaction Levels, etc. The future work also entails including not just TC metrics but also adaptation behaviors such as opening/closing windows and modifying clothing, as prediction ``tasks'' in the multi-task model. 

\section{Author Contributions}
Author contributions using the Contributor Roles Taxonomy (CRediT) \cite{credit} are highlighted below
\begin{enumerate}
    \item Betty Lala: Conceptualization, Methodology, Investigation, Visualization, Data Curation, Validation (Data), Formal analysis (Data), Writing - Original Draft, Writing - Review \& Editing
    \item Hamada Rizk: Methodology, Software, Investigation (Software), Formal analysis (Software), Validation (Software), Visualization, Writing - Original Draft
    \item Srikant Manas Kala: Methodology, Data Curation, Validation (Data), Investigation (Data), Formal analysis (Data), Visualization, Writing - Original Draft, Writing - Review \& Editing, Project administration
    \item Aya Hagishima: Supervision, Conceptualization, Methodology, Writing - Review \& Editing, Project administration, Resources, Funding acquisition
\end{enumerate}
\section{Acknowledgment}
The research was funded by the Sasakawa Scientific Research Grant of the Japan Science Society. 

Authors thank the principals, teachers, and most importantly, the students of the following reputed schools in Dehradun, India, for participating in this study: Grace Academy, Cambrian Hall, Kendriya Vidyalaya Salawala, St. Thomas College, \& Jaswant Modern Senior Secondary School.

Authors are grateful to Mrs. Pushpa Manas, Director of School Education (Retd.), State of Uttarakhand, India, for facilitating this study.

\section{Methods and Informed Consent}
The survey was conducted in the 5 schools with permission and informed consent from the school administration. Confidentiality and privacy of student data is maintained by removing all indicators such as the name of the School, student, or teacher from the data used in predictive modeling.

% \subsection{Performance on Ashre data}
% \subsection{transfer learning our data}

%\appendix

% If you have bibdatabase file and want bibtex to generate the
\color{black}
 \bibliographystyle{IEEEtran} 
 \bibliography{ref, MLTCR2}
 \vspace{4cm}
\begin{IEEEbiography}[{\includegraphics[width=1in,height=1.25in,clip,keepaspectratio]{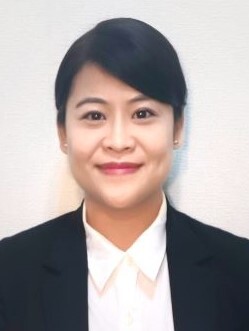}}]{B\MakeLowercase{etty} L\MakeLowercase{ala}} is a doctoral researcher at the Interdisciplinary Graduate School of Engineering Sciences (IGSES), Kyushu University, Japan. Prior to this, she was an Assistant Professor in the Faculty of Architecture, Manipal University, India. She was also a professional Architect for 3 years specializing in villas and luxury high-rise buildings. She received her masters degree in Architecture from IIT Roorkee, India. She was a recipient of the DAAD scholarship, and carried out her masters' research at TU Berlin, Germany. She also received the prestigious Sasakawa scientific research grant in the year 2020-21. Her interests lie in the field of thermal comfort, research survey methodology, data analysis, urban design, and UX design.
\end{IEEEbiography}
\begin{IEEEbiography}[{\includegraphics[width=1in,height=1.25in,clip,keepaspectratio]{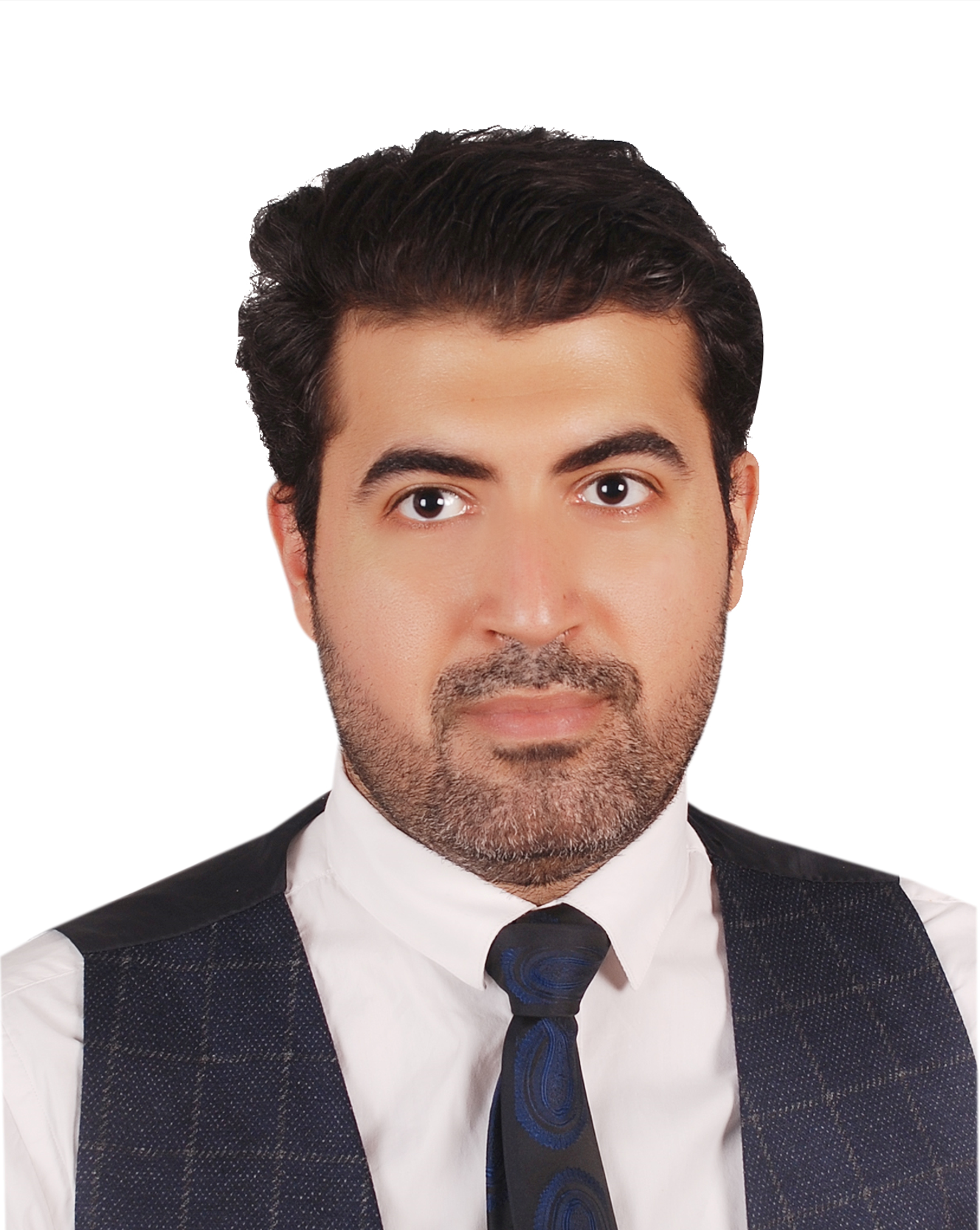}}] {D\MakeLowercase{r}. H\MakeLowercase{amada} R\MakeLowercase{izk}} received the B.E. and M.E. degrees in Computer and Automatic Control engineering from Tanta University and the Ph.D. degree in Computer Science and Engineering from E-JUST in 2020. He is currently an Assistant Professor at Tanta University, Egypt, and Osaka University, Japan. He has been working in mobile and pervasive computing, spatial intelligence, and applied machine learning research areas. He has been involved in several projects funded by many academic and industrial organizations such as NTRA Egypt, Uber, USA, ASTEP JST, Kakenhi, JSPS, Japan... etc.  He has authored several publications in top journals and conferences and holds a number of patents. Hamada has also received several personal awards such as the silver medal in the 4th ACM SigSpatial competition held in Chicago, 2019 and the same year honored as an outstanding young researcher and received the Romberg award by the HLF foundation, Germany, among other awards.
\end{IEEEbiography}
\begin{IEEEbiography}[{\includegraphics[width=1in,height=1.25in,clip,keepaspectratio]{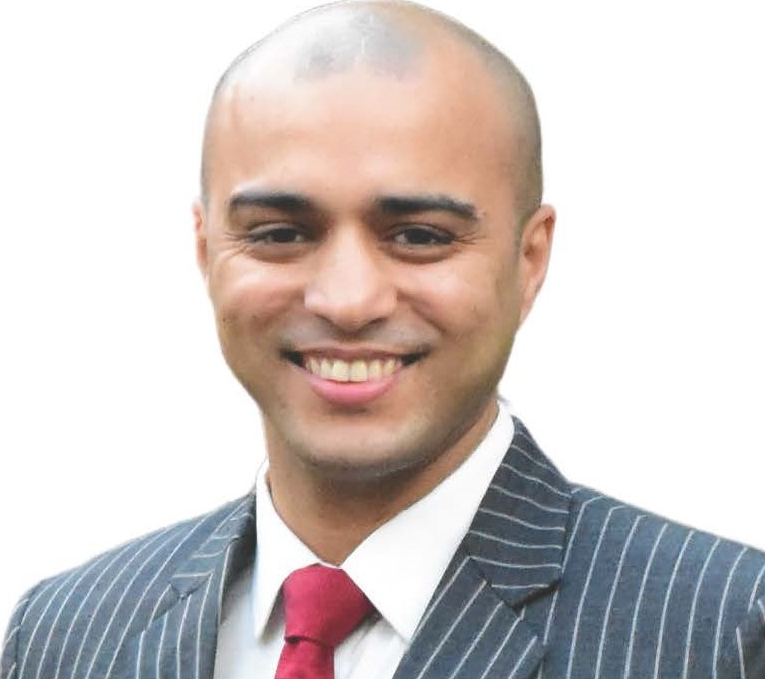}}]{S\MakeLowercase{rikant}  M\MakeLowercase{anas} K\MakeLowercase{ala}} is a doctoral researcher at the Mobile Computing Lab, Osaka University, Japan. He received his M.Tech degree in Computer Science and Engineering from  IIT Hyderabad, India. He has been awarded the Employee Excellence Award by Infosys and IIT Hyderabad Research Excellence Award in 2016 and 2017. He led his startup team to the semifinals of Ericcson Innovation Awards 2020 and the Impact Summit of Hult Prize 2021. His interests lie in the domain of Extended Reality, applied AI/ML, Venture Capital investment analysis, and Unlicensed and 5G Networks,
\end{IEEEbiography}
\begin{IEEEbiography}[{\includegraphics[width=1in,height=1.25in,clip,keepaspectratio]{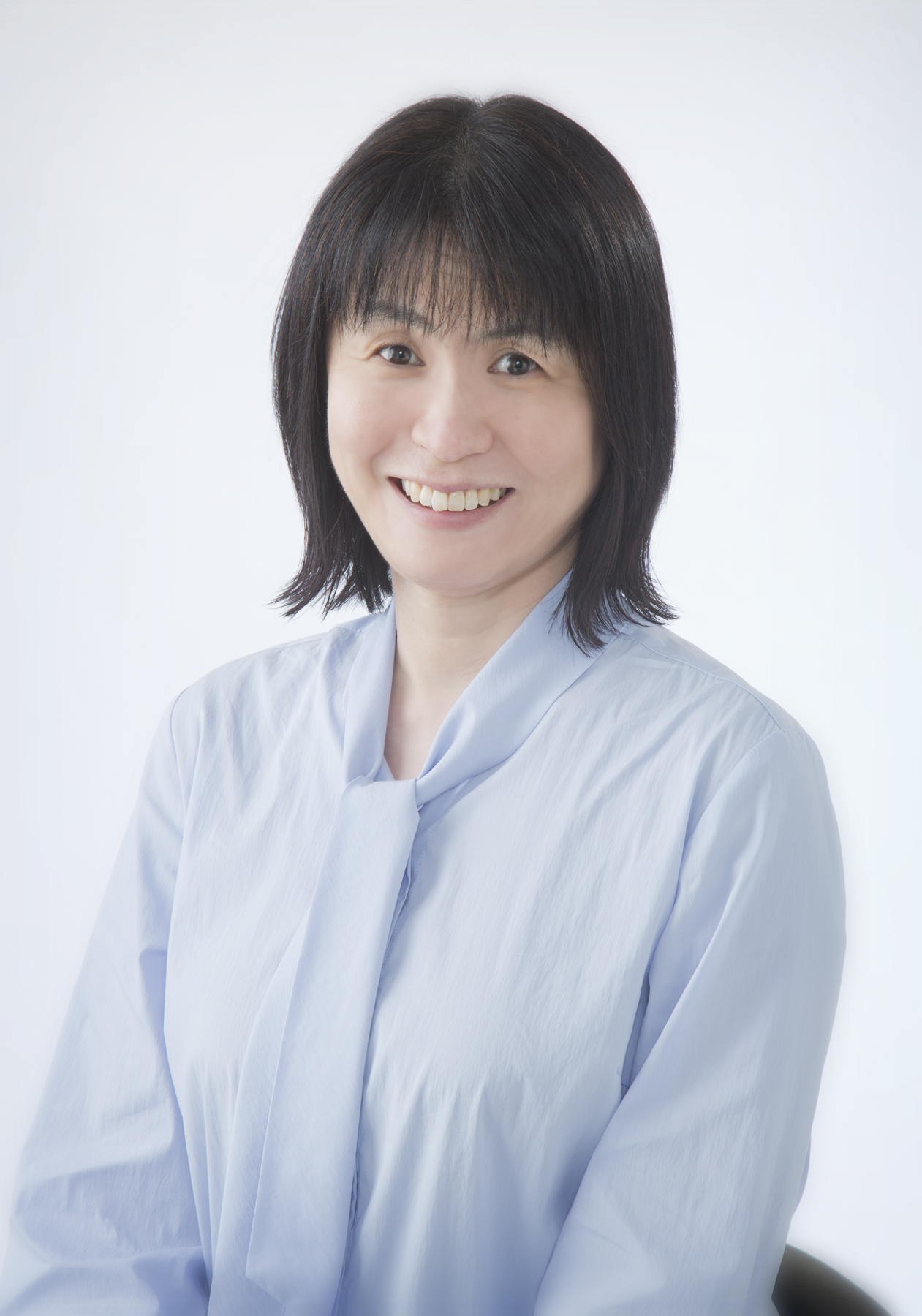}}]{D\MakeLowercase{r}. A\MakeLowercase{ya} H\MakeLowercase{agishima}} received the Dr. Eng degree from the Interdisciplinary Graduate School of Engineering Sciences (IGSES), Kyushu University, Fukuoka, Japan, in 2005. She is currently a Professor in the Faculty of Engineering Sciences, Kyushu University. She has worked mainly in the research areas of urban climatology and building environment.  Her current research interests include zero-energy buildings/houses, energy-related occupants behaviours, and sustainable built environment.
\end{IEEEbiography}
\end{document}